%% file: main.tex
\documentclass[10pt,twocolumn,letterpaper,table,xcdraw,dvipsnames]{article}

\usepackage{cvpr}              % To produce the CAMERA-READY version
\usepackage{graphicx}
\usepackage{esvect}
\usepackage{booktabs}
\usepackage{multirow}

\usepackage{tabularx}
\usepackage{ifthen}

\newcommand{\connor}{\textcolor{blue}}

\usepackage{derivative}
\usepackage{amsmath,amssymb}
\usepackage{bm}
\usepackage[symbol]{footmisc}
\usepackage{comment}
\usepackage{algorithm}
\usepackage{algpseudocode}

\let\amssymbboxplus\boxplus
\let\amssymbboxminus\boxminus

\renewcommand{\boxplus}{\mathbin{\mathop\amssymbboxplus}}
\renewcommand{\boxminus}{\mathbin{\mathop\amssymbboxminus}}

\newcommand{\state}{\mathbb{S}}
\newcommand{\real}{\mathbb{R}}
\newcommand{\ut}{\mathbf{u}^t}
\newcommand{\quat}{\mathbf{q}}

\newcommand{\pix}{\mathbf{x}}

\newcommand{\event}{\mathbf{e}_i}
\newcommand{\sworld}{\mathbf{s}}

\newcommand{\attitude}{\quat}


\definecolor{cvprblue}{rgb}{0.21,0.49,0.74}
\usepackage[pagebackref,breaklinks,colorlinks,citecolor=cvprblue]{hyperref}

\def\paperID{5817} % *** Enter the Paper ID here
\def\confName{CVPR}
\def\confYear{2025}

\title{EBS-EKF: Accurate and High Frequency Event-based Star Tracking}

\author{Albert W. Reed\thanks{Kitware}, Connor Hashemi\footnotemark[1], Dennis Melamed\footnotemark[1], Nitesh Menon, Keigo Hirakawa\thanks{University of Dayton}, and Scott McCloskey\\
Kitware\\
{\tt\small connor.hashemi@kitware.com}\\
University of Dayton\footnotemark[2]\\
{\tt\small keigo.hirakawa@udayton.edu}
}

\newcommand{\loadcamerareadysections}{0}

\begin{document}

\twocolumn[{%
\renewcommand\twocolumn[1][]{#1}%
\maketitle
\begin{center}
    \centering
    \captionsetup{type=figure}
    \includegraphics[width=1\textwidth, page=2]{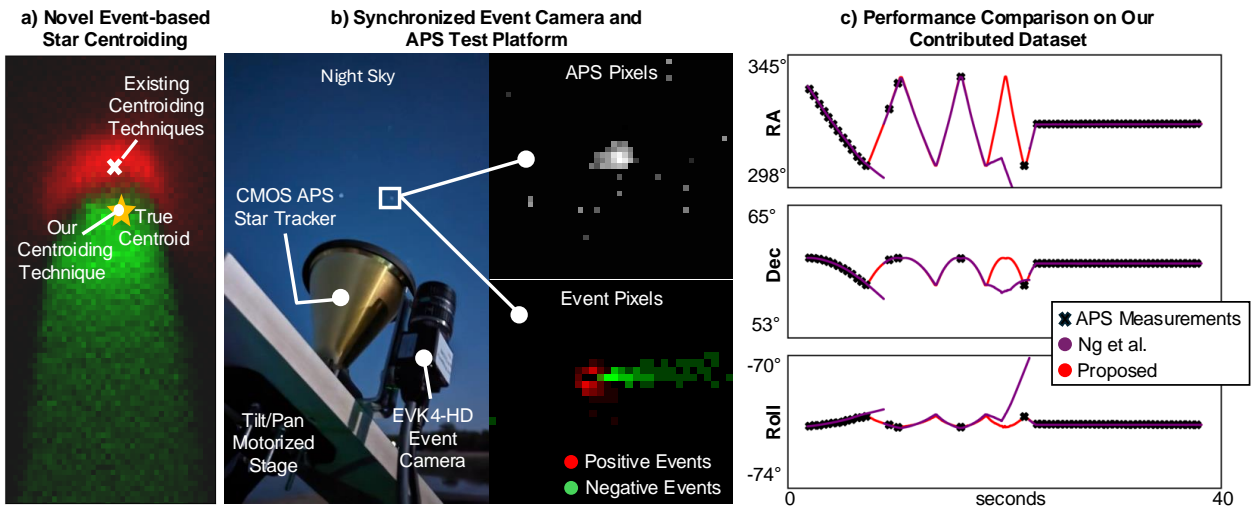}
    \captionof{figure}{We present EBS-EKF, an event-based star tracking algorithm that combines a novel centroiding technique with an extended Kalman filter and is validated using the first ground-truthed dataset of event streams from real stars. (a) Our centroiding technique accounts for event camera behavior in low light, enabling more accurate tracking. (b) We develop a data collection setup of a EVK4-HD event camera that is rigidly mounted and synchronized with a space-ready APS star tracker. The insets show APS (top) and event camera (bottom) pixels for 8 Cygni, a subgiant star in the constellation Cygnus. (c) We demonstrate that our attitude estimates (red) are more accurate than existing methods (purple) ~\cite{Ng2022}, and that we can operate above the 3 deg/sec cutoff of the APS star tracker, highlighting the utility of our method for high-frequency star tracking.}

    %, which is representative of a conventional star tracking sensor. 
    %\captionof{figure}{{\bf Overview of paper}. (a) data collection rig to capture event-based and intensity-based star tracking data. (b) examples captured by the commercial star tracker and the event-based camera. (c) the paper's novel star event model, which predicts a changing event profile based on the star's brightness. (d) results of the proposed EVT-EKF method, which has a much faster update rate than the commercial star tracker and is more accurate than previous event-based methods. \connor{Sub real data in for b,d}}
    \label{fig:teaser}
\end{center}%
}]

\footnotetext[1]{Denotes equal contribution.}

\ifthenelse{\equal{\loadcamerareadysections}{0}}%
  {
        \input{new_new_secs/0_abstract}

\input{new_new_secs/1_intro}

        \input{new_new_secs/2_related_work}

        \input{new_new_secs/3_methods}
        \input{new_new_secs/5_results.tex}

        \input{new_new_secs/6_conclusions}

  }% If part
  {% Else part

\input{new_secs/0_abstract}\input{new_secs/1_intro}\input{new_secs/2_related_work}
        \input{new_secs/3_methods_single_event.tex}
        \input{new_secs/4_experiments}

\input{new_secs/5_results}\input{new_secs/6_conclusions}
 }
{
    \small
    \bibliographystyle{ieeenat_fullname}
    \bibliography{main}
}

\clearpage

\input{new_new_secs/X_suppl}


\end{document}

%% file: new_new_secs/0_abstract.tex
\begin{abstract}
%\vspace{-28pt}

Event-based sensors (EBS) are a promising new technology for star tracking due to their low latency and power efficiency, but prior work has thus far been evaluated exclusively in simulation with simplified signal models.  
We propose a novel algorithm for event-based star tracking, grounded in an analysis of the EBS circuit and an extended Kalman filter (EKF). 
We quantitatively evaluate our method using real night sky data, comparing its results with those from a space-ready active-pixel sensor (APS) star tracker. 
We demonstrate that our method is an order-of-magnitude more accurate than existing methods due to improved signal modeling and state estimation, while providing more frequent updates and greater motion tolerance than conventional APS trackers. 
We provide all code\footnote{\text{https://gitlab.kitware.com/nest-public/kw\_ebs\_star\_tracking\#}} and the first dataset of events synchronized with APS solutions.

%but their performance is currently limited by over-simplified signal models. 
%but their performance on real-world data remains unverified.

%we develop new centroiding theory for accurately estimating the position of stars in event camera pixel space. We then design and implement a new event-based star tracking method called EBS-EKF which uses a 3D extended Kalman filter to track the camera's attitude with accuracy down to 30 arcseconds and updates rates up to 1 kHz. We collect a dataset of the night sky by synchronizing event camera outputs with those from a commercial star tracker, allowing us to quantify the accuracy of our approach. We demonstrate through our experiments that our proposed method outperforms existing event-based star tracking algorithms. 
\end{abstract}

%Star trackers perform high-precision attitude tracking of satellites using optical sensors, producing right ascension, declination, and roll measurements in support of spacecraft control. 

% unexplored, and all previous methods have only been tested in simulation.

%% file: new_new_secs/1_intro.tex
% \usepackage{multirow}
\begin{table*}[]
\caption{Overview of related works. Color indicates high (green), medium (yellow) and low (red) performance.  ``s'' denotes seconds.}
\setlength{\tabcolsep}{3.5pt} % default 6
\renewcommand{\arraystretch}{1.25} % Default value: 1
\scriptsize
\begin{tabular}{|
>{\columncolor[HTML]{EFEFEF}}l |
>{\columncolor[HTML]{FFFFFF}}l 
>{\columncolor[HTML]{FFFFFF}}l |
>{\columncolor[HTML]{FFCCC9}}l 
>{\columncolor[HTML]{FFFC9E}}l lll|}
\hline
\multicolumn{1}{|c|}{\cellcolor[HTML]{C0C0C0}}                                                                                              & \multicolumn{2}{c|}{\cellcolor[HTML]{C0C0C0}\textbf{Method Design}}                                                                                                                                                                                       & \multicolumn{5}{c|}{\cellcolor[HTML]{C0C0C0}\textbf{Key Features}}                                                                                                                                                                                                                                                                                                                                                                                                                                                                                                                                \\ \cline{2-8} 
\multicolumn{1}{|c|}{\multirow{-2}{*}{\cellcolor[HTML]{C0C0C0}\textbf{\begin{tabular}[c]{@{}c@{}}Star Tracking\\  Algorithm\end{tabular}}}} & \multicolumn{1}{c|}{\cellcolor[HTML]{EFEFEF}\begin{tabular}[c]{@{}c@{}}State \\ Estimation\end{tabular}}                                      & \multicolumn{1}{c|}{\cellcolor[HTML]{EFEFEF}\begin{tabular}[c]{@{}c@{}}Centroiding\\ Method\end{tabular}} & \multicolumn{1}{c|}{\cellcolor[HTML]{EFEFEF}\begin{tabular}[c]{@{}c@{}}Quantitative\\ Evaluation\end{tabular}}           & \multicolumn{1}{c|}{\cellcolor[HTML]{EFEFEF}Camera State}                                                                 & \multicolumn{1}{c|}{\cellcolor[HTML]{EFEFEF}\begin{tabular}[c]{@{}c@{}}Absolute \\ Orientation Updates\end{tabular}}   & \multicolumn{1}{c|}{\cellcolor[HTML]{EFEFEF}\begin{tabular}[c]{@{}c@{}}Centroiding\\ Accuracy\end{tabular}} & \multicolumn{1}{c|}{\cellcolor[HTML]{EFEFEF}\begin{tabular}[c]{@{}c@{}}Update\\  Rate\end{tabular}} \\ \hline
{\color[HTML]{333333} Chin et al. \cite{chin_star_2019}}                                                                                                          & \multicolumn{1}{l|}{\cellcolor[HTML]{FFFFFF}\begin{tabular}[c]{@{}l@{}}iterative closest point (ICP)\end{tabular}} & mean of all events                                                                                        & \multicolumn{1}{l|}{\cellcolor[HTML]{FFCCC9}LCD screen}                                                                  & \multicolumn{1}{l|}{\cellcolor[HTML]{FFFC9E}3D rotation}                                                                  & \multicolumn{1}{l|}{\cellcolor[HTML]{FFCCC9}\begin{tabular}[c]{@{}l@{}}required\\ often (e.g., every 5 s)\end{tabular}}                  & \multicolumn{1}{l|}{\cellcolor[HTML]{FFFC9E}$\sim$1.8 pixels}                                               & \cellcolor[HTML]{FFCCC9}20-100 Hz                                                                   \\ \hline
{\color[HTML]{333333} Bagchi et al. \cite{bagchi_event-based_2020}}                                                                                                        & \multicolumn{1}{l|}{\cellcolor[HTML]{FFFFFF}\begin{tabular}[c]{@{}l@{}}multi-resolution \\ hough transform (HT)\end{tabular}}                 & mean of positive events                                                                                   & \multicolumn{1}{l|}{\cellcolor[HTML]{FFCCC9}LCD screen}                                                                  & \multicolumn{1}{l|}{\cellcolor[HTML]{FFFC9E}3D rotation}                                                                  & \multicolumn{1}{l|}{\cellcolor[HTML]{FFCCC9}\begin{tabular}[c]{@{}l@{}}required \\ often (e.g., every 5 s)\end{tabular}}                 & \multicolumn{1}{l|}{\cellcolor[HTML]{FFCCC9}$\sim$2.7 pixels}                                               & \cellcolor[HTML]{FFCCC9}20-100 Hz                                                                   \\ \hline
{\color[HTML]{333333} \begin{tabular}[c]{@{}l@{}}Ng et al. \cite{Ng2022} \end{tabular}}                                                & \multicolumn{1}{l|}{\cellcolor[HTML]{FFFFFF}\begin{tabular}[c]{@{}l@{}}2D Kalman filter \\ (2D-KF)\end{tabular}}                              & \begin{tabular}[c]{@{}l@{}}high-pass \\ filtered intensity\end{tabular}                           & \multicolumn{1}{l|}{\cellcolor[HTML]{FFCCC9}LCD screen}                                                                  & \multicolumn{1}{l|}{\cellcolor[HTML]{FFFC9E}\begin{tabular}[c]{@{}l@{}}2D position\\ + 2D velocity\end{tabular}}          & \multicolumn{1}{l|}{\cellcolor[HTML]{FFFC9E}\begin{tabular}[c]{@{}l@{}}required\\ intermittently (e.g., every 30 s)\end{tabular}}         & \multicolumn{1}{l|}{\cellcolor[HTML]{FFCCC9}$\sim$3.0 pixels}                                               & \cellcolor[HTML]{9AFF99}500-1000 Hz                                                                 \\ \hline
{\color[HTML]{333333} \textbf{Ours}}                                                                                                        & \multicolumn{1}{l|}{\cellcolor[HTML]{FFFFFF}\textbf{\begin{tabular}[c]{@{}l@{}}3D extended Kalman filter \\ (EBS-EKF)\end{tabular}}}          & \textbf{\begin{tabular}[c]{@{}l@{}}mean of positive events \\ with magnitude offset\end{tabular}}         & \multicolumn{1}{l|}{\cellcolor[HTML]{9AFF99}\textbf{\begin{tabular}[c]{@{}l@{}}night sky and\\ LCD screen\end{tabular}}} & \multicolumn{1}{l|}{\cellcolor[HTML]{9AFF99}\textbf{\begin{tabular}[c]{@{}l@{}}3D rotation\\ + 3D velocity\end{tabular}}} & \multicolumn{1}{l|}{\cellcolor[HTML]{9AFF99}\textbf{\begin{tabular}[c]{@{}l@{}}only on \\ initialization\end{tabular}}} & \multicolumn{1}{l|}{\cellcolor[HTML]{9AFF99}\textbf{$\sim$0.4 pixels}}                                      & \cellcolor[HTML]{9AFF99}\textbf{500-1000 Hz}                                                        \\ \hline
\end{tabular}
\label{tab:related:methods}
\end{table*}

% and\\  Latif et al. \cite{latif_high_2023}

\section{Introduction}
\label{sec:intro}
 
Star trackers estimate the precise attitude --- right ascension (RA), declination (Dec), and roll --- of a camera with respect to the celestial coordinate frame, making them a standard sensor for space navigation. Compared to other sensors, star trackers provide highly accurate measurements, with attitude estimate errors on the order of arcseconds (1 arcsecond = 1/3600 degrees). Modern trackers generally employ active-pixel sensor (APS) cameras to identify star centroids and determine camera attitude. While APS trackers perform well when stationary, their finite exposure times and frame processing overhead usually limit update rates to 2-10 Hz, constraining performance during rapid attitude adjustments~\cite{liebe_accuracy_2002}.

% typically using active-pixel sensor (APS) cameras that image measure stars with a fixed exposure time. 

% which pose challenges for high velocity maneuvers. Specifically, star trackers provide relatively slow update rates, typically operating at 2-10 Hz, 

%

Event-based cameras have recently emerged as a promising technology for enhancing star tracker performance. Also known as event-based sensors (EBS), event cameras mimic human vision by capturing microsecond-resolution ``events" in each pixel, indicating whether the light reaching a pixel has increased or decreased by a specified threshold. Unlike APS cameras, EBS measurements are asynchronous and can reach temporal resolutions down to the microsecond, suggesting that event-based star trackers could achieve significantly faster update rates. Furthermore, because events are triggered only by changes in the scene, most pixels remain inactive when imaging stars in the night sky, which could lead to lower power consumption. 

Given these potential advantages, several recent studies have investigated the utility of EBS-based star tracking~\cite{bagchi_event-based_2020, chin_star_2019, Ng2022, latif_high_2023}. While these studies have introduced new algorithms, many questions remain before EBS star trackers can potentially replace APS trackers. First, all quantitative evaluations thus far have been conducted in simulation, where star fields are displayed on a screen. To date, the performance of EBS cameras on real-world star fields, where noise and low-light imaging pose significant challenges, has not been quantified. Second, existing techniques from literature yield considerable variability in EBS star centroiding, leading to suboptimal attitude estimation. Finally, no real-world EBS datasets or open-source EBS trackers are currently available, making it challenging to benchmark and improve existing methods.

In this work, we present a novel event-based star centroiding technique by modeling the dynamics of a star imaged with an EBS circuit. Leveraging this model, we design an event-based star tracker based on an extended Kalman filter (EKF). We evaluate our algorithm in both lab settings and real night sky experiments. Our specific contributions are as follows:

\begin{enumerate} 
    \item A novel method for event-based star centroiding, based on rigorous analysis of event camera circuit behavior in low-light conditions. 
    
    \item An extended Kalman filter tailored for EBS-based star tracking, providing 3D attitude and velocity measurements with update rates up to 1 KHz and enhanced accuracy.
 
    \item The first dataset of events paired with APS tracker quaternion solutions, collected from real starfields, enabling quantitative evaluation of our method against an APS tracker and existing EBS techniques.

    \item Our code and dataset are released as open-source to support reproducibility of research.
\end{enumerate}

%% file: new_new_secs/2_related_work.tex
\section{Related Work}
\label{sec:related-work}

%We begin by discussing the background of conventional star tracking, followed by recent developments in event-based star tracking. Lastly, we review relevant works on low-light event-based imaging, which form the theoretical basis for our centroiding technique.

%\subsection{EBS Imaging}
%\label{sec:related-work:lowlightdvs}
\noindent \textbf{EBS Imaging:} An EBS detects changes in pixel brightness by generating measurement tuples termed ``events''. Events indicate when a pixel's intensity rises or falls past a set threshold; each event includes the pixel's location, a microsecond timestamp, and polarity (i.e. ``positive'' if intensity increased and ``negative'' if it decreased) \cite{lichtsteiner_128_2008, gallego_event-based_2020}.  Compared to APS, EBS offers lower power consumption, faster pixel response, and a larger dynamic range. EBS imaging has proven useful for many applications, including autonomous driving \cite{gehrig_low-latency_2024}, robotics \cite{wang_ev-catcher_2022}, high-FPS cameras \cite{kodama_122m_2023, scheerlinck_continuous-time_2018}, wavefront imaging \cite{grose_convolutional_2024}, and photogrammetry \cite{yu_eventps_2024}.

\label{sec:dvs-trackers}

% We provide a brief overview on the operation of APS star trackers, highlighting relevant works related to the one we employ as a point of reference for our DVS star tracker. 

\noindent \textbf{APS Star Tracking: } In APS tracking, stationary stars are typically modeled as 2D Gaussians on the imaging plane, then localized using center-of-gravity (COG) \cite{stone_comparison_1989, akondi_improved_2010} or Gaussian fitting \cite{delabie_accurate_2014, wan_star_2018} techniques. Once stars are localized on the image plane, various techniques can translate these positions into attitude estimates. A common approach, and the one employed by our reference APS tracker, estimates an \textit{absolute} rotation by cross-referencing identified stars with a star catalog (we use \textit{Astrometry.net}~\cite{lang_astrometry_2010}). Other methods derive relative rotations from the frame-to-frame optical flow~\cite{li_extended_2017, Sun_of_kalman}. 

%\todo{Describe what absolute measurements are. }

%While DVS represents a different sensing modality, it shares the same pinhole camera geometry of an APS sensor, and many of the existing event-based tracking techniques are borrowed directly from APS star tracking. 

%The two primary components of a star tracker are the centroiding of stars on the imaging plane and the algorithms responsible for converting centroids into an attitude. As summarized in Figure~\ref{fig:related:dvs_table}, several studies have explored methods for event-based star tracking.

%Star trackers typically use state estimation techniques, such as Kalman Filters, to smooth attitudes estimated from noisy centroids. 
 % While event cameras share pinhole camera geometry of APS cameras, the fundamental differences between the sensing modalities, including measuring changes in brightness instead of raw intensity, require EBS-specific tracking algorithms. 

%While this method provided proof-of-concept for EBS star tracking, it proposed a \textit{synchronous} approach that did not exploit the \textit{asynchronous} nature of EBS measurements. Noting this limitation, the

\noindent \textbf{EBS Star Tracking:} Chin et al.~\cite{chin_star_2019} were the first to address event-based star tracking, proposing a method that integrates events into synchronous frames and uses the iterative closest point (ICP) algorithm to estimate frame-to-frame rotations; we shorthand the algorithm from~\cite{chin_star_2019} as ``ICP'' throughout this paper. They observed that this technique is prone to noisy attitude estimates and drift, necessitating periodic absolute measurements from a reference star catalog~\cite{lang_astrometry_2010}. The authors proposed a follow-up work that used the Hough transform to asynchronously update the tracker rotation~\cite{bagchi_event-based_2020} from streaks in the event stream --- we refer to this method as ``Hough''. While this approach demonstrated asynchronous EBS star tracking, it achieved higher error rates compared to ICP~\cite{chin_star_2019}. 

More recent works propose an asynchronous 2D Kalman filter for EBS star tracking~\cite{Ng2022, latif_high_2023}, which we call ``2D KF'' throughout this paper. 
The Kalman filter models star position and velocity as 2D linear translations and ignores the non-linear effects of roll. 
Their simulation results demonstrated superior update rates and accuracy to prior works~\cite{chin_star_2019, bagchi_event-based_2020}.

The asynchronous Kalman filter works are most related to ours, but we introduce several key advancements. First, previous works either implicitly~\cite{chin_star_2019, bagchi_event-based_2020,latif_high_2023} or explicitly~\cite{Ng2022} model the likelihood of events using a Gaussian distribution centered on the star, whereas we model it as a Gaussian whose offset from the star's location depends on the star's intensity. Secondly, we employ an asynchronous extended Kalman filter that estimates 3D rotations and angular velocities, and demonstrate that 3D modeling is essential for effective tracking on real data. 

As summarized in Table~\ref{tab:related:methods}, existing methods quantify tracking performance on simulated starfields. 
Ng et al.~\cite{Ng2022} demonstrate an ability to track stars in the night sky but do not quantify tracking performance due to lack of a ground truth. 
Our work is the first to quantify the performance of event-based tracking algorithms on real data.

\input{graphics/tables/dataset_algorithm_table}
\begin{figure}[t]
  \centering
   \includegraphics[width=1\linewidth]{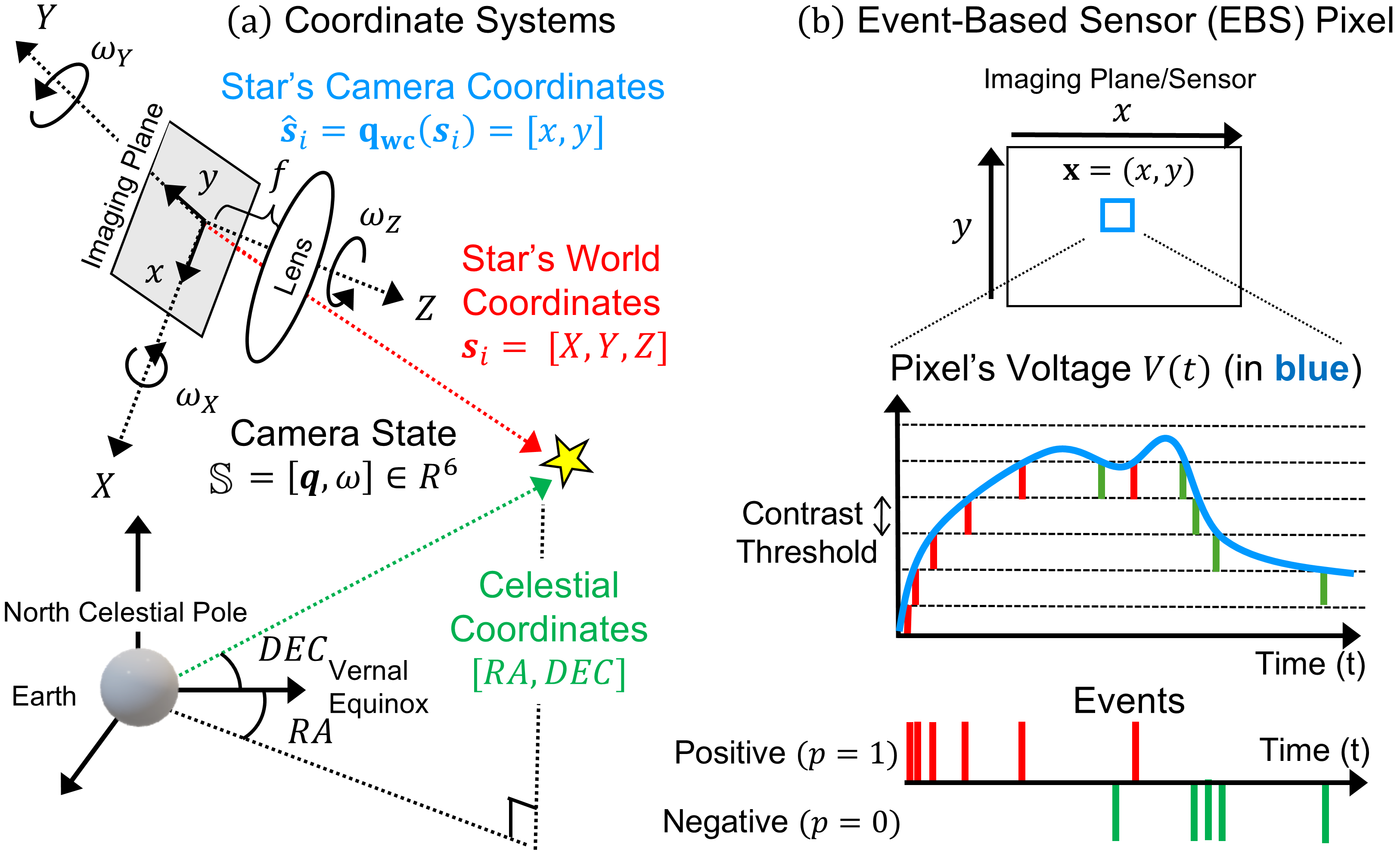}
   \caption{Measurement model for event-based star tracking. (a) depicts the coordinate systems used; (b) depicts an example operation of an EBS pixel as a star's projection passes over it.}
   \label{fig:methods:meas_model}
\end{figure}

 % In this example the star's velocity is 50 pixels/s. 
% This highlights the importance of considering illumination conditions to accurately centroid stars from the event distribution. 

\begin{figure*}[t]
  \centering
   \includegraphics[width=1\linewidth]{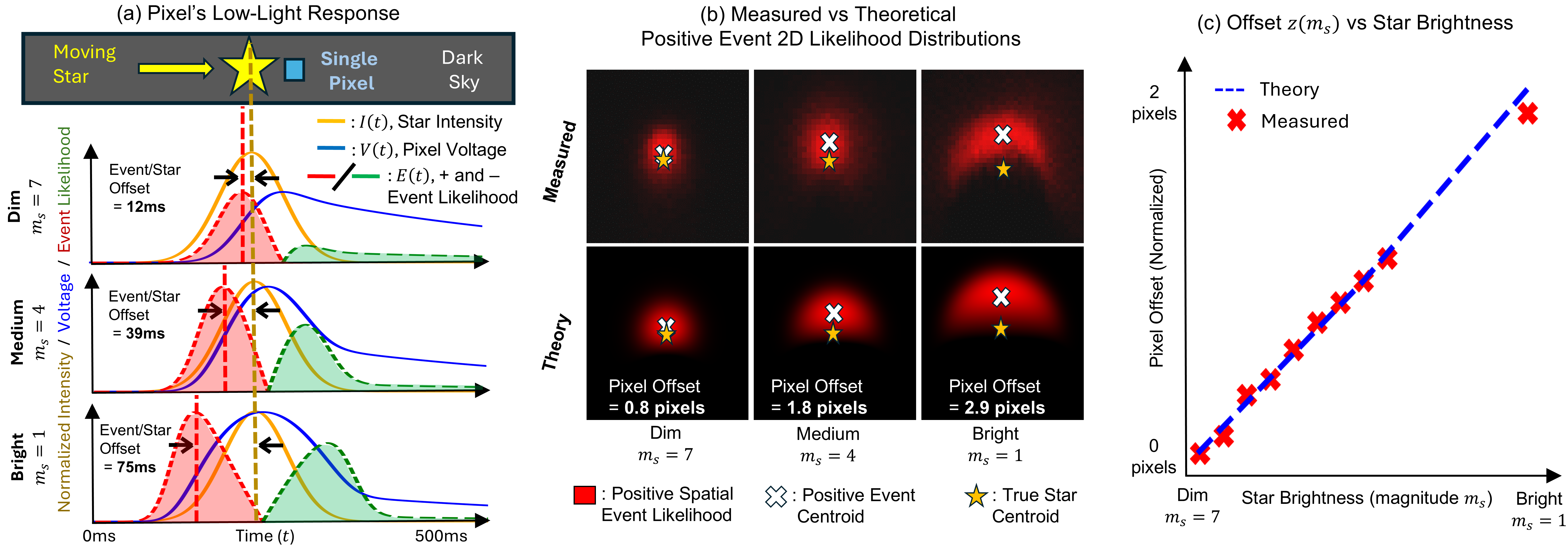}
   \caption{In low-light conditions, a star's brightness introduces a variable delay between the observed peak of positive events and its true locations. (a) depicts the numerical solution of event likelihood $E_{\text{LL}}(t)$ in Eq \ref{eq:methods:diffeq} compared to star intensity $I(t)$ for a single pixel. Bright stars produce a distribution of positive and negative events that follows the derivative of the intensity. Dim stars, however, yield a distribution of positive events that peaks nearer to the peak intensity, and the negative events fall to the tail of the distribution. (b) compares the theoretical spatial positive event likelihood distribution (calculated by solving Eq \ref{eq:methods:complete_spatial_likelihood}) to those measured in our night sky dataset. (c) depicts the offset $z(m_s)$ between the centroid of the positive events and the star's true centroid as a function of the star's brightness (i.e. relative magnitude $m_s$). The offsets are normalized to have a minimum offset of 0.}
   \label{fig:methods:event_profile}
\end{figure*}
%the theoretical event distributions are calculated by solving Eq \ref{eq:methods:complete_spatial_likelihood} for multiple star intensity profiles based on the pixel position and star speed as done in Eq \ref{eq:methods:pixel_solve}.

%% file: new_new_secs/3_methods.tex
\section{EBS-EKF Method}
\label{sec:methods}
We first describe our problem setting and camera geometry. We then define a signal model that relates the likelihood of events to star positions. We use this signal model in conjunction with an extended Kalman filter to perform high-frequency and accurate event-based star tracking.

\subsection{Measurement Geometry}
\label{subsec:measurement-geometry}

As illustrated in Fig.~\ref{fig:methods:meas_model}, we aim to use EBS measurements of stars to estimate the camera state $\state = [\quat, \mathbf{\omega}]$, where $\quat \in SO(3)$ is a quaternion in the 3D rotation group, and $\mathbf{\omega} \in \real^3$ denotes the 3D angular velocity. We measure the set of $N$ stars $\mathbf{S} \in \real^{N \times 3}$; each star is 3D point normalized to the unit sphere with negligible parallax effects~\cite{forbes2015fundamentals}. Each star has an apparent magnitude $m_{s_i}$, which is a measure of the star brightness on a reverse logarithmic scale (i.e., smaller values indicate brighter stars)\footnote{This works considers the set of stars with apparent magnitude $\leq 7$.}. A single star $\mathbf{s}_i \in \mathbf{S}$ with 3D coordinates $(X, Y, Z)$ is rotated into the camera's frame using a world-to-camera quaternion, $\hat{\mathbf{s}}_i = \attitude_{\text{wc}}(\mathbf{s}_i)$\footnote{We typically omit the wc subscript for brevity.}, where $\attitude_{\text{wc}}(\cdot)$ denotes the quaternion's rotation of a 3D vector. After rotating, the star is projected onto the 2D imaging plane using the pinhole camera model:
\begin{align}
    x = f\frac{X}{Z}, \;\; \ y = f\frac{Y}{Z}, \label{eq:measurement-fn-2}
\end{align}
where $f$ is the focal length. The EBS indicates a change of intensity at a pixel by triggering an event $\event = (\mathbf{x},p,t)$, where $\mathbf{x} = (x, y)$ denotes the pixel, $p$ denotes the event polarity, and $t$ denotes the timestamp. 

\noindent \textbf{Celestial Coordinates:} As illustrated in Figure~\ref{fig:methods:meas_model}, a camera's attitude can be defined as the optical axis' orientation to the celestial coordinate frame, in degrees of right-ascension (RA), declination (Dec), and roll (about the z-axis). In alignment with star tracking literature, we use \textit{across} orientation (i.e., pointing direction) to refer to Ra and Dec, and \textit{about} to refer to roll orientation.

% The objective of this work is to accurately centroid stars using events and then use these centroids to asynchronously update the camera state. 

%Accurate estimation of the camera's state $\state$ requires precise localization of star centroids in the imaging plane. To solve this, w

%\subsection{Star Event Model}
%\label{sec:event-signal-model}

%and the physical principles of the DVS photoreceptor in low-light scenarios.

\subsection{Event Likelihood Model}

\textbf{Continuous-time Event Likelihood:} We now derive the likelihood of an event triggering based on a star's brightness and dynamics on the imaging plane. 
A model for EBS operation is shown in Fig \ref{fig:methods:meas_model}b, where the EBS pixel's photoreceptor converts irradiance to pixel voltage and triggers events~\cite{lichtsteiner_128_2008}.
%In a simplified version of the DVS, the DVS pixel's photoreceptor converts image intensity to pixel voltage.  This simplified DVS operation is shown 

In this work, we define the \textit{event likelihood} $E(t)$, which is a model that defines the event rate of a EBS pixel $\pix$ at each time $t$. Because there are two polarities of events, $E(t)>0$ defines the likelihood of a positive event while $E(t)<0$ defines the likelihood of a negative event. In bright (i.e. $\geq 100$ lux) conditions, the event likelihood can be modeled as the continuous change in the photoreceptor's voltage over time, or
\begin{equation}
    E(t) = \frac{dV(t)}{dt} = \frac{d}{dt}\text{log}(I(t)/I_0 + 1),
    \label{eq:methods:eventrate}
\end{equation}
where $V(t)$ is the pixel's voltage, $I(t)$ is the pixel's image intensity, and $I_0$ is an intensity offset defined by the circuit and the dark current \cite{graca_shining_2023}. %The dark current $I_0$ is typically ignored in  we will explain becomes relevant in our low-light imaging setting. 

%$I_0$ is typically negligible in bright conditions and therefore most models assume $\text{log}(I(t)/I_0 + 1) \propto \text{log}(I(t))$.

%Examples of this are illustrated Figure~\ref{fig:methods:event_profile} (a)., where we plot event likelihoods for different brightness conditions.

Following existing star imaging optics, the star's point-spread function on the sensor can be well-approximated by a multivariate, isotropic Gaussian ~\cite{delabie_accurate_2014}. Therefore, we model the star intensity $I(t)$ at pixel location $\mathbf{x}$ as
\begin{equation}
    I(t) \! = \! \frac{1}{\sqrt{2 \pi}}|\Sigma|^{-\frac{1}{2}} \text{exp} \Big (-\frac{1}{2}(\mathbf{x}_0-\mathbf{v}t )\Sigma^{-1}(\mathbf{x}_0-\mathbf{v}t )^{T}\Big ),
   \label{eq:method:gauss_star}
\end{equation}
where $\mathbf{x}_0 = (x_0, y_0)$ is the star's offset from the pixel $\pix$ at $t=0$, $\mathbf{v} = (v_x, v_y)$ is the star's image-plane velocity, and $\Sigma = \sigma^2_s \mathbf{I}_{2x2}$, where $\sigma^2_s$ is the star's variance across pixels. Plugging the star intensity of Eq.~\ref{eq:method:gauss_star} into the event rate of Eq.~\ref{eq:methods:eventrate} defines the measurement model leveraged by existing works~\cite{Ng2022}, where likelihood $E(t)$ is derived as the derivative of the Gaussian intensity. 

% , which we account for by modeling the intensity-dependent bandwidth of the DVS pixels

% which can be explained by low-light EBS circuit phenomenology

However, \textit{\textbf{our key insight is that the event likelihood's position and shape depends on star brightness}}, which can be explained by low-light EBS circuit phenomenology. In low-light, the bandwidth of the EBS circuit is proportional to the photocurrent, making the pixel's voltage $V(t)$ behave like a single-order low-pass filter with an intensity-dependent cutoff frequency $f_c$~\cite{lichtsteiner_128_2008, posch_qvga_2011, liu_seeing_2024, Hu_Liu_Delbruck_2021}. We incorporate this into a low-light (LL) event likelihood model using the first-order differential equation
\begin{equation}
    \begin{aligned}
        &E_{\text{LL}}(t) = \frac{dV(t)}{dt} =  2\pi \cdot f_c(\Tilde{I}(t)) \cdot \left[\Tilde{I}(t) - V(t)\right],
        \label{eq:methods:diffeq}
    \end{aligned}
\end{equation}
where $\Tilde{I}(t)=\text{log}(I(t)/I_0 + 1)$ is the pixel's photocurrent and $f_c(\Tilde{I}(t))$ is the intensity-dependent cutoff frequency \cite{posch_qvga_2011, Hu_Liu_Delbruck_2021}.
We model the bandwidth dependence as a linear function of photocurrent $\Tilde{I}(t)$:
\begin{equation}
    f_c(\Tilde{I}(t)) \approx b + a\Tilde{I}(t)  ,
    \label{eq:methods:fc_equation}
\end{equation}
where $b$ is the cutoff frequency when the pixel is dark (i.e. $\Tilde{I}(t) = 0$) and $a$ is its linear dependence on photocurrent $\Tilde{I}(t)$. Intuitively, the lower bandwidth associated with dim stars delays the peak of the event likelihood compared to the higher bandwidth of bright stars. Since the event likelihood follows the derivative of a Gaussian, this implies that the likelihood for bright stars generally leads the true centroid, while the likelihood for dim stars lags (or coincides) with it. We observe that this linear model facilitates accurate prediction of the pixel offsets observed in our real night sky data, as shown in the agreement of theory and measured offset curves in Figure \ref{fig:methods:event_profile}c.

As shown in Figure~\ref{fig:methods:event_profile}a, the negative event likelihood exhibits high entropy for dim stars. Considering their significant latency in low-light and substantial processing overhead, we choose to process only positive events.

\noindent \textbf{Spatial Event Likelihood:} Using our continuous-time event likelihood, we can express the likelihood of a positive event $e_i=(\mathbf{x}_i,t_i)$ as:
\begin{equation}
\mathcal{L}(e_i | \mathbf{x_0}, \mathbf{v}) = \text{max} \big( 0,E_{\text{LL}}(t_i-(\pix_i-\pix_0)/\mathbf{v}) \big ),
\label{eq:methods:complete_spatial_likelihood}
\end{equation}
given a star with constant linear velocity $\mathbf{v}$ and position $\mathbf{x_0}$.\footnote{The max operator effectively ignores negative events by ensuring the likelihood remains positive.} Figure~\ref{fig:methods:event_profile}b illustrates several example spatial likelihoods, where the positive event mean is offset from the true centroid depending on the star brightness. Figure~\ref{fig:methods:event_profile}c plots this offset, and we observe agreement with real data when numerically solving our model with choice parameters of the dark current $I_0$, cutoff-frequency slope $a$, and offset $b$. We note the presence of a ``bowshock'' effect for bright stars (also observed by \cite{benson_simulation_2022}), where the edges of the event likelihood are more delayed than the center --- this is due to a lower cutoff frequency in the dimmer regions of the star, which induces more delay relative to the bright center.

\noindent \textbf{Gaussian Approximation:} We aim to use the likelihood in Eq.~\ref{eq:methods:complete_spatial_likelihood} for real-time estimation of the camera attitude, but it is a first-order differential equation that is computationally expensive to solve. Instead of using it directly, we propose approximating it with an isotropic Gaussian distribution:
\begin{equation}
\mathcal{L}(e_i | \mathbf{x_0}, \mathbf{v}) \sim \mathcal{N} \Big (\pix_i -  \big[\pix_0-\mathbf{\bar{v}} \cdot z(m_s) \big ]  ,     \sigma^2_s \mathbf{I}_{2x2} \Big ),
\label{eq:methods:gauss_approx_likelihood}
\end{equation}
where the function $z: \real \rightarrow \real$ corrects for the star's magnitude-dependent $m_s$ offset (see Figure~\ref{fig:methods:event_profile}c) and $\mathbf{\bar{v}}$ denotes the normalized linear velocity. We quantify the accuracy of this approximation in Section~\ref{sec:centroiding-results}.

% \connor{Because we utilize small time batches ($\sim$1ms), the approximation ignores the distance traveled by the star in the time batch (i.e. $t \mathbf{v} \approx 0$).} 

%, this approximation the derived likelihood while requiring significantly less computation.

% $\sigma_s$ denotes the standard deviation of the distribution, 

\begin{algorithm}
\caption{EBS-EKF Star Tracking}\label{alg:cap}
\begin{algorithmic}
\State \textbf{Input: } Positive event stream $\mathcal{E} = \{\mathbf{x}, t\}_{i=0:N}$, 
\State \textbf{Output: } Camera State $\state$ (3D rotation and angular velocity)

\State \textbf{Initialize:} $\state_0 \gets \text{astrometry~\cite{lang_astrometry_2010} with binned events}$

%\State $k \gets 0$
\State $\text{r} \gets$ \text{max search radius}

\For{each event $e_i = (\mathbf{x}_i, t_i)$}

    \State $\hat{\state}_i \gets \text{EKF predict at $t_i$}$ 

    \State $\mathbf{x}_s \gets \text{closest projected catalog star to } \mathbf{x}_i$
    \If{distance($\mathbf{x}_i$, $\mathbf{x}_s$) $\leq$ r}
        \State $\bar{\mathbf{v}}_s \gets$ \text{star's normalized linear velocity via $\hat{\state}_i$}
        \State $m_s \gets$ \text{star's apparent magnitude}
        \State $\hat{\mathbf{x}}_i \gets \mathbf{x}_i + \bar{\mathbf{v}}_s \cdot z(m_s)$ \# apply offset correction
        \State $\state_i \gets \text{EKF update with $\hat{\mathbf{x}}_i$}$
    \EndIf
\EndFor
\end{algorithmic}
\label{alg:tracking-alg}
\end{algorithm}

\subsection{State Estimation and Tracking Algorithm}

We now use the event likelihood in a Bayesian estimation framework to estimate the camera state. Our measurements are the set of $N$ positive events $\mathcal{E}_N = \{(\mathbf{x}_i, t_i)\}_{i=0:N}$, where we denote each event as $\event \in \mathcal{E}$. Under a Markov assumption, the probability distribution over all states and measurements is,

\vspace{-20pt}
\begin{equation}
p(\state_0, \! ...\state_N; e_0, \! ...e_N) \! = \!p(\state_0) \! \prod_{i=0}^{N} \! \mathcal{L}(e_i | \state_i)p(\state_i | \state_{i - 1}). \! \!  
\label{eq:likelihood}
\end{equation}
\vspace{-10pt}

\noindent Intuitively, this formulation estimates the states as an interpolation of the measurement likelihood $\mathcal{L}(e_i | \state_i)$ and prior $p(\state_i | \state_{i - 1})$ on the state's evolution. Notably, we define the measurement likelihood $\mathcal{L}(e_i | \state_i)$ as conditioned on the state $\state$; the state defines the star position $\mathbf{x}_0$ and velocity $\mathbf{v}$ shown in Eq.~\ref{eq:methods:gauss_approx_likelihood}. 

\noindent \textbf{Extended Kalman Filter:} We propose using an EKF to recursively estimate the posterior distribution $p(\state_i | \mathcal{E}_i)$ derived from Eq.~\ref{eq:likelihood}, which is the state conditioned on the measurements up to the current time step~\cite{barker1995bayesian}\footnote{We provide more details on Eq.~\ref{eq:likelihood}'s Bayesian formulation of the Kalman filter in the supplementary material.}. The EKF estimates the state through \textit{predict} and \textit{update} phases, which require key matrices as input: the forward model Jacobian $\mathbf{F}$, processing noise $\mathbf{Q}$, and measurement matrix $\mathbf{H}$. In the supplemental material, we derive the measurement matrix based on the projection geometry of Section~\ref{subsec:measurement-geometry} and the forward model and process noise matrices for a general constant velocity model in 3D space. We find that the EKF enables more accurate state estimation compared to other methods (see results) and recognize the potential of many other state estimation techniques (e.g., ~\cite{wan_unscented_2000, dellaert2017factor}) as avenues for future research.

%for which we provide more detial in the supplementary material. Specifically, we use an extended Kalman filter (EKF)~\cite{julier1997new} to capture the non-linear dynamics of our state, which   We use a \textit{constant velocity} prior, which smooths state predictions and mitigates the negative impacts of noisy measurements. 

\noindent \textbf{Full Tracking Algorithm: }Alg.~\ref{alg:tracking-alg} outlines our algorithm for estimating the camera state from events, for which we provide more details in the supplemental material. We initialize the camera attitude by binning events with a small time window (e.g., 30 ms), identifying event clusters with DBSCAN clustering~\cite{ester_density-based_nodate}, and passing the cluster centroids to astrometry~\cite{lang_astrometry_2010} to determine an initial attitude. 
We then use events to update the state. Specifically, we use the event's timestamp to update the EKF prediction, and check if the event is within the radius of a catalog star, given the current attitude. If so, we use the star's magnitude and linear velocity to apply our intensity-dependent offset correction, and then use it to update the camera state using our EKF update equations (see supplemental). We note that this algorithm is asynchronous and operable from single positive events, facilitating the 1 KHz update rate used in this work. 

%% file: new_new_secs/5_results.tex
\section{Results}
\label{sec:centroiding-results}

We now discuss validation of our event-based centroiding theory and star tracking algorithm.  

\subsection{Laboratory Centroiding Experiments}
\label{sec:monitor_exps}
Our first set of experiments are performed in a laboratory setting where we can manually configure star brightness, position, and velocity. We use these experiments to quantify the accuracy of centroiding methods and to evaluate the accuracy of our Gaussian approximation (Eq.~\ref{eq:methods:gauss_approx_likelihood}) for the analytical likelihood (Eq.~\ref{eq:methods:diffeq}).
\begin{figure}
  \centering
   \includegraphics[width=1\linewidth]{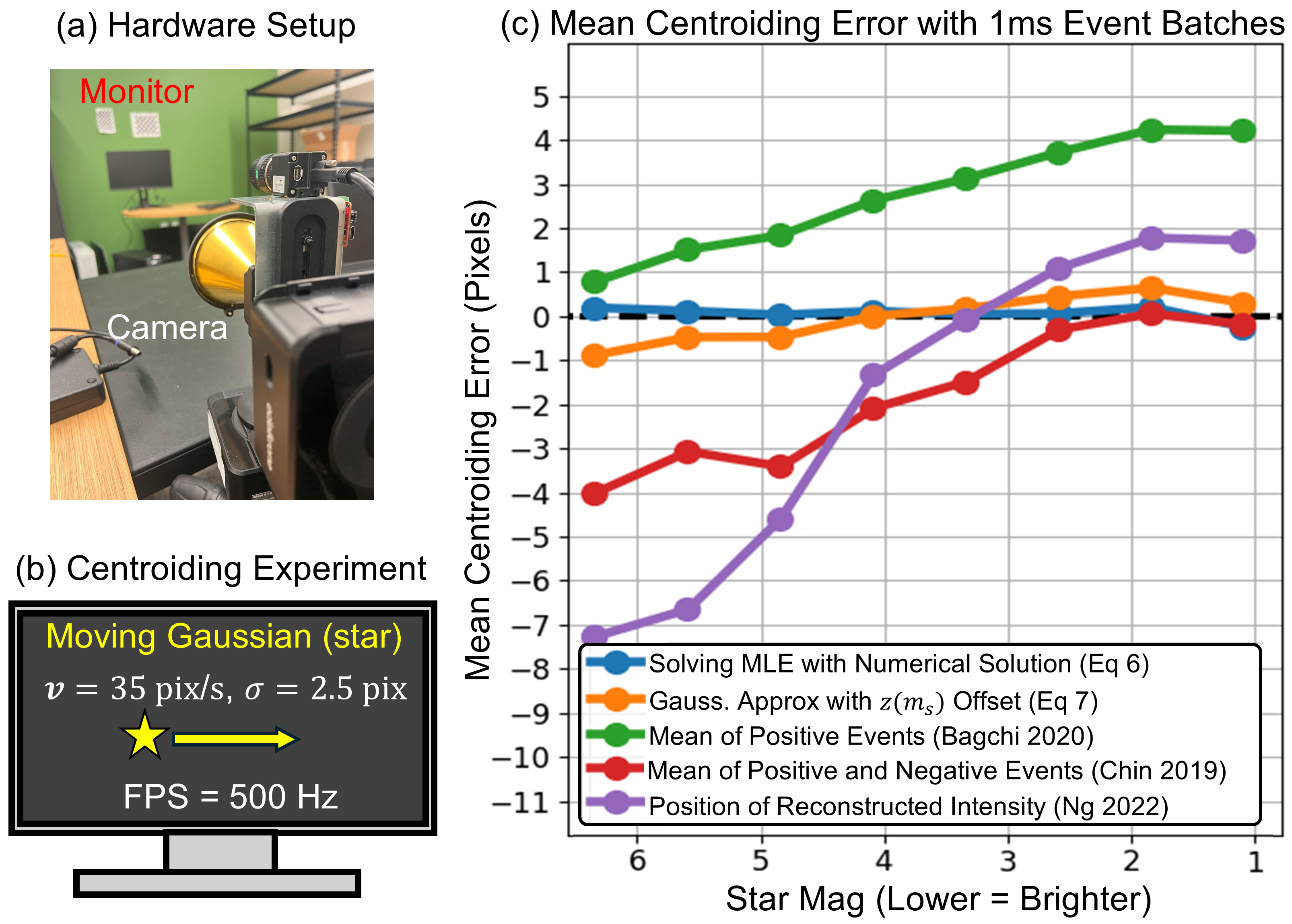}
    \caption{Testing the centroiding accuracy of Eqs \ref{eq:methods:complete_spatial_likelihood} and \ref{eq:methods:gauss_approx_likelihood} via an LCD monitor. (a) depicts the hardware setup. (b) depicts the centroiding experiment on the high-update rate monitor, where the star is moving at 35 pixels per second and has a 2.5 pixel standard deviation to emulate stars in our night sky dataset. (c) depicts the centroiding accuracies of our methods (blue and orange curve) compared to those of previous works. The black dotted line depicts zero centroiding error. Details about the synchronization and other aspects are included in supplementary. }
   \label{fig:results:monitor_centroiding}
    \vspace{-10pt}
\end{figure}
% We change the brightness of the star by placing neutral-density filters over the camera.

\noindent \textbf{Implementation:} Our experimental setup shown in Fig \ref{fig:results:monitor_centroiding}(a-b) consists of an EVK4-HD event camera focused on a high-refresh rate monitor. We use the monitor to simulate stars by displaying gamma-corrected, isotropic Gaussians that move at a constant speed. We set the apparent magnitude of the stars on the monitor with a combination of placing neutral-density (ND) filters in front of the camera and adjusting the brightness of display pixels. We use the transformation matrix between camera and monitor pixels to compute an error between estimated and ground truth star positions. We provide more implementation details in the supplemental.

% via a calibration pattern. Because star position on the monitor and the transformation matrix are known,
%\todo{Through a combination of pixel intensity and ND filters we emulate star's with apparent magnitude 1-7. We provide more details. We provide more details in the supplemental. }

\noindent \textbf{Results: }
Fig. \ref{fig:results:monitor_centroiding}c displays the centroiding results for all methods against different star brightness levels. We use 1 ms batches to estimate the centroid using: (1) maximum likelihood estimation (MLE) of our analytical model in Eq.~\ref{eq:methods:complete_spatial_likelihood} (blue), (2) our Gaussian approximation with intensity-dependent offset of Eq \ref{eq:methods:gauss_approx_likelihood} (orange), (3) the mean of the positive events~\cite{bagchi_event-based_2020} (green), (4) the mean of positive and negative events~\cite{chin_star_2019,latif_high_2023} (red), and (5) the position of the reconstructed intensity~\cite{Ng2022} (purple). Our analytical model is the most accurate of the set (error of $0.16\pm0.18$  pixels), but requires numerically solving the differential equation in Eq \ref{eq:methods:diffeq}. Our Gaussian approximation has an error of $0.42\pm0.48$ pixels, and is orders of magnitudes faster to calculate, taking less than 0.1ms per batch compared to over 100ms for the analytical model. Both of our methods outperform previous approaches in terms of centroiding accuracy and consistency, as they account for intensity-dependent centroiding effects and low-light circuitry effects, which prior methods do not.

\begin{figure}
    \centering
    \includegraphics[width=\linewidth]{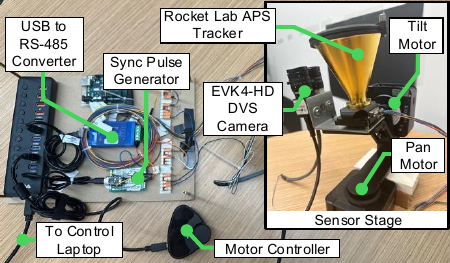}
    \caption{Our night sky data collection system with key components labeled. The sync pulse generator ensures that we can temporally correlate the Rocket Lab APS measurements with the EVK4-HD event camera measurements. The Pan and Tilt Motors are used to sweep both star trackers across the sky based on pre-defined motion patterns.
    }
    \label{fig:experimental:data_collect_rig}
\end{figure}

\subsection{Real World Star Tracking Experiments}
\label{subsec:realworldresults}

We also quantify the performance of ours and existing methods on event data captured from stars in the night sky.  

\noindent \textbf{Hardware Configuration:} 
As illustrated in Figure~\ref{fig:experimental:data_collect_rig}, we time-synchronize the event streams from a Prophesee EVK4-HD sensor with the outputs of a space-ready APS star tracker (Rocket Lab ST-16RT2) via a synchronization pulse generator as their fields of view are moved across the night sky. 
The EVK4-HD sensor is fitted with a 35mm lens, yielding a field of view (FOV) of 10.2\textdegree x5.7\textdegree. 
The ST-16RT2 is factory-installed with a 16mm lens, yielding a FOV of 20\textdegree x15\textdegree. 
We compute a relative rotation between the rigidly co-mounted cameras for use in our quantitative evaluation. 
Specifically, we can use the relative rotation to compute a difference between APS and EBS tracks as $||\mathbf{q}_{\text{APS}} \cdot \mathbf{q}_{\text{relative}} - \mathbf{q}_{\text{\text{EBS}}}||_2^2$, where $\mathbf{q}$ denotes a quaternion, the dot operator denotes quaternion multiplication, and the minus operator denotes the difference between quaternions. 
We provide additional details on obtaining their relative rotations in the supplemental material. 
We also provide details on the cameras' parameters and our method for obtaining the offset correction curve plotted in Fig.~\ref{fig:methods:event_profile} (c).

%\todo{FIX ground truth error margins sentence. Move report accuracy to the bottom of the paragraph.}
 
\noindent \textbf{Metrics and Dataset:} We report the mean squared difference between the EBS and APS tracks in both across (pointing) and about (roll) accuracies, as is standard in literature~\cite{liebe_accuracy_2002}. Performance is measured relative to solutions provided from the Rocket Lab tracker, which has its own error margins. Specifically, the Rocket Lab tracker is rated for 5 arcseconds across and 50 arcseconds about at slew rates up to 3 deg/sec in yaw and pitch. As such, we speak of methods' performances in terms of their \textit{difference} with respect to the APS tracker. We report differences in arcseconds (denoted using '').   

%The reported metrics should be interpreted as a test of each method's performance with respect to the reference solutions provided by the Rocket Lab tracker.

\begin{figure}
   \centering
   \includegraphics[width=\linewidth]{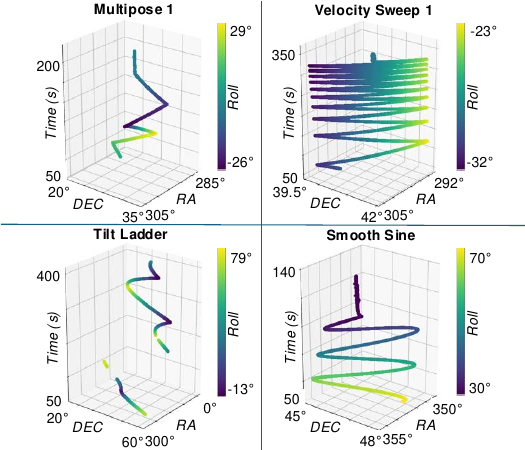}
   \caption{Plots of four real night-sky tracks that were estimated with our EBS-EKF algorithm. Right Ascension and Declination are shown along the bottom, and time flows vertically. Roll is indicated by the color of the line.}
   \label{fig:experimental:track_profiles}
\end{figure}

This paper tests all methods on 14 real star data tracks total, which were collected from multiple locations in North America. Track durations are between 1 and 5 minutes, and were acquired using various camera trajectories, velocities, and regions of the night sky. % (0.18 deg/sec - 1.8 deg/sec)
Visualizations of four tracks (estimated with our tracking algorithm) are shown in Figure~\ref{fig:experimental:track_profiles}. We summarize key dataset details in the top portion of Table~\ref{table:main-results} and provide additional details in the supplementary material.

%We obtain the offset calibration curve (shown in Figure~\ref{fig:methods:event_profile}) on a hold-out track to mitigate data contamination. 

\noindent \textbf{Method Implementations:} We implemented all methods in Python and ran experiments on an M2 Silicon MacBook. Open-source code was not available for existing methods (2D-KF~\cite{Ng2022}, ICP~\cite{chin_star_2019}, and Hough~\cite{bagchi_event-based_2020}), so we implemented it to the best of our abilities and performed parameter sweeps to optimize the performance on our dataset. Unlike our method, existing approaches require frequent absolute measurements to stay on track. Absolute measurements are obtained by cross-referencing centroids with a star catalog (in our case, using \textit{Astrometry.net}~\cite{lang_astrometry_2010}). While they are typically too slow for real-time processing, they are required by existing methods~\cite{chin_star_2019, bagchi_event-based_2020}, and can be computed offline. Thus, in the spirit of maximizing their performance, we chose to provide frequent (i.e., every 5 seconds) precomputed absolute measurements to ICP and Hough and to 2D KF when it lost track. 

\input{graphics/tables/results_table}

\begin{figure}
  \centering
\includegraphics[width=1\columnwidth]{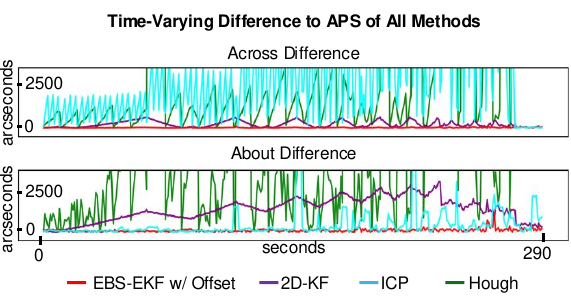}
   \caption{Real World Velocity Sweep 1. Differences of all methods with APS reference measurements. Our method obtains significantly lower difference with the APS reference measurements. Existing methods drift off track by hundreds of arcseconds due to inaccurate centroiding and erroneous state estimation.}
   \label{fig:results:track-errors}
    \vspace{-10pt}
\end{figure}

\noindent \textbf{Results: } We summarize the performance of all methods on our night-sky tracks in Table~\ref{table:main-results}. We observe that our method is typically within 100 arcseconds (across and about) of the APS tracker, whereas existing methods deviate by up to 1-2 degrees (i.e., several thousand arcseconds).  2D-KF is the most competitive to ours, but periodically drifts off track due its 2D state estimation failing to model the 3D dynamics of real data. Notably, ICP is more accurate than Hough, in agreement with reported results ~\cite{bagchi_event-based_2020}. Figure~\ref{fig:results:track-errors} plots the time-varying difference for all methods, showing that existing methods generate saw-tooth shaped difference plots --- they drift off track and accumulate difference, but are then snapped on track by the periodic absolute measurements. 

\begin{figure}
  \centering
\includegraphics[width=.99\linewidth, page=1]{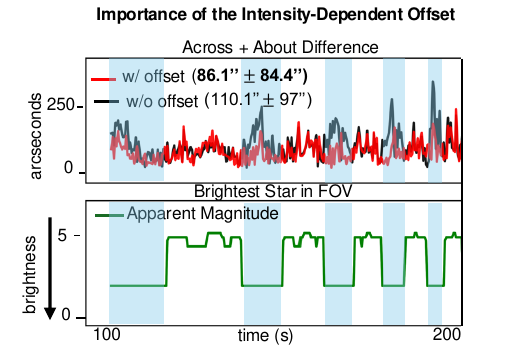}
   \caption{Snippet from Velocity Track 1. Without offset correction (black), error spikes when a bright star (magnitude 2.23) enters the field of view (shaded blue). Our offset correction accounts for the star's brightness, attenuating these spikes and lowering the average difference by over 20 arcseconds.}
   \label{fig:results:bright-star-in-fov}
   \vspace{-10pt}
\end{figure}

The bottom rows of Table~\ref{table:main-results} summarize our method's performance with and without using the intensity-dependent centroiding offset. We observe that it typically improves performance by 10-20 arcseconds, usually when a bright star is within the FOV. Since there is 1-2 pixel centroid offset between our brightest and dimmer stars (see Fig.~\ref{fig:methods:event_profile}c), and the FOV of each event pixel is approximately 30 arcseconds, improvements within these ranges agree with intuition. We provide an example scenario from Velocity Sweep 1 of how a bright star can drastically affect EBS tracking performance in Fig.~\ref{fig:results:bright-star-in-fov}, showing that attitude estimation error spikes without using our offset correction.  

%In Figure~\ref{fig:teaser}(c), we show our method compared with 2D-KF~\cite{Ng2022}, showing that both maintain track better than the APS tracker at speeds up to 7.5 deg/sec. When APS measurements are provided, we are an order of magnitude closer to them compared with the 2D-KF.

\noindent \textbf{High Velocity Track:} Figure~\ref{fig:teaser}c compares our method with 2D-KF on a 7.5 deg/sec velocity track, which is not included in Table~\ref{table:main-results} since the APS tracker fails to provide consistent solutions past the velocity of 3 deg/sec. Both ours and Ng et al.'s~\cite{Ng2022} method produce high frequency estimates of the tracks, but only ours reconstructs the correct track, and it maintains an order of magnitude closer agreement to the available APS measurements (total difference of 80.4 arcseconds versus 774.3 arcseconds, respectively). We provide a video of this experiment in the supplemental material.

%% file: graphics/tables/results_table.tex
\begin{table*}[h]
\centering
\footnotesize
\begin{tabular}{l|lllllllllllll|}
\cline{2-14}
 & \multicolumn{13}{c|}{\cellcolor[HTML]{9B9B9B}\textbf{Track Name}} \\ \cline{2-14} 
 & \multicolumn{7}{c|}{\textbf{Velocity Sweep}} & \multicolumn{4}{c|}{\textbf{Multipose}} & \multicolumn{1}{c|}{} & \multicolumn{1}{c|}{} \\
 & \multicolumn{1}{c|}{1} & \multicolumn{1}{c|}{2} & \multicolumn{1}{c|}{3} & \multicolumn{1}{c|}{4} & \multicolumn{1}{c|}{5} & \multicolumn{1}{c|}{6} & \multicolumn{1}{c|}{7} & \multicolumn{1}{c|}{1} & \multicolumn{1}{c|}{2} & \multicolumn{1}{c|}{3} & \multicolumn{1}{c|}{4} & \multicolumn{1}{c|}{\multirow{-2}{*}{\textbf{\begin{tabular}[c]{@{}c@{}}Tilt \\ Ladder\end{tabular}}}} & \multicolumn{1}{c|}{\multirow{-2}{*}{\textbf{\begin{tabular}[c]{@{}c@{}}Smooth \\ Sine\end{tabular}}}} \\ \cline{2-14} 
 & \multicolumn{13}{c|}{\cellcolor[HTML]{9B9B9B}\textbf{Metadata}} \\ \hline
\textbf{Duration (s)} & \multicolumn{1}{l|}{\cellcolor[HTML]{EFEFEF}290} & \multicolumn{1}{l|}{\cellcolor[HTML]{EFEFEF}290} & \multicolumn{1}{l|}{\cellcolor[HTML]{EFEFEF}167} & \multicolumn{1}{l|}{\cellcolor[HTML]{EFEFEF}160} & \multicolumn{1}{l|}{\cellcolor[HTML]{EFEFEF}160} & \multicolumn{1}{l|}{\cellcolor[HTML]{EFEFEF}160} & \multicolumn{1}{l|}{\cellcolor[HTML]{EFEFEF}150} & \multicolumn{1}{l|}{\cellcolor[HTML]{EFEFEF}151} & \multicolumn{1}{l|}{\cellcolor[HTML]{EFEFEF}150} & \multicolumn{1}{l|}{\cellcolor[HTML]{EFEFEF}145} & \multicolumn{1}{l|}{\cellcolor[HTML]{EFEFEF}265} & \multicolumn{1}{l|}{\cellcolor[HTML]{EFEFEF}342} & \cellcolor[HTML]{EFEFEF}85 \\ \hline
\textbf{\begin{tabular}[c]{@{}l@{}}RA / Dec / Roll \\ Range (\textdegree)\end{tabular}} & \multicolumn{1}{l|}{\cellcolor[HTML]{EFEFEF}\begin{tabular}[c]{@{}l@{}}13.9\\ 2.3\\ 8.9\end{tabular}} & \multicolumn{1}{l|}{\cellcolor[HTML]{EFEFEF}\begin{tabular}[c]{@{}l@{}}9.8\\ 6.6\\ 14.4\end{tabular}} & \multicolumn{1}{l|}{\cellcolor[HTML]{EFEFEF}\begin{tabular}[c]{@{}l@{}}6.8\\ 2.9\\ 0.8\end{tabular}} & \multicolumn{1}{l|}{\cellcolor[HTML]{EFEFEF}\begin{tabular}[c]{@{}l@{}}5.7\\ 3.8\\ 0.9\end{tabular}} & \multicolumn{1}{l|}{\cellcolor[HTML]{EFEFEF}\begin{tabular}[c]{@{}l@{}}3.6\\ 4.7\\ 1.1\end{tabular}} & \multicolumn{1}{l|}{\cellcolor[HTML]{EFEFEF}\begin{tabular}[c]{@{}l@{}}1.3\\ 5.2\\ 1.3\end{tabular}} & \multicolumn{1}{l|}{\cellcolor[HTML]{EFEFEF}\begin{tabular}[c]{@{}l@{}}6.4\\ 3.2\\ 5.4\end{tabular}} & \multicolumn{1}{l|}{\cellcolor[HTML]{EFEFEF}\begin{tabular}[c]{@{}l@{}}21.7\\ 17.2\\ 55.2\end{tabular}} & \multicolumn{1}{l|}{\cellcolor[HTML]{EFEFEF}\begin{tabular}[c]{@{}l@{}}28.01\\ 18.6\\ 36.0\end{tabular}} & \multicolumn{1}{l|}{\cellcolor[HTML]{EFEFEF}\begin{tabular}[c]{@{}l@{}}7.1\\ 2.7\\ 44.9\end{tabular}} & \multicolumn{1}{l|}{\cellcolor[HTML]{EFEFEF}\begin{tabular}[c]{@{}l@{}}11.3\\ 8.1\\ 90.2\end{tabular}} & \multicolumn{1}{l|}{\cellcolor[HTML]{EFEFEF}\begin{tabular}[c]{@{}l@{}}0.01\\ 44.4\\ 92.8\end{tabular}} & \cellcolor[HTML]{EFEFEF}\begin{tabular}[c]{@{}l@{}}5.1\\ 3.8\\ 40.3\end{tabular} \\ \hline
\textbf{Max Velocity (\textdegree/s)} & \multicolumn{1}{l|}{\cellcolor[HTML]{EFEFEF}1.8} & \multicolumn{1}{l|}{\cellcolor[HTML]{EFEFEF}1.8} & \multicolumn{1}{l|}{\cellcolor[HTML]{EFEFEF}1.8} & \multicolumn{1}{l|}{\cellcolor[HTML]{EFEFEF}1.8} & \multicolumn{1}{l|}{\cellcolor[HTML]{EFEFEF}1.8} & \multicolumn{1}{l|}{\cellcolor[HTML]{EFEFEF}1.6} & \multicolumn{1}{l|}{\cellcolor[HTML]{EFEFEF}1.7} & \multicolumn{1}{l|}{\cellcolor[HTML]{EFEFEF}0.9} & \multicolumn{1}{l|}{\cellcolor[HTML]{EFEFEF}0.8} & \multicolumn{1}{l|}{\cellcolor[HTML]{EFEFEF}0.8} & \multicolumn{1}{l|}{\cellcolor[HTML]{EFEFEF}0.8} & \multicolumn{1}{l|}{\cellcolor[HTML]{EFEFEF}0.6} & \cellcolor[HTML]{EFEFEF}0.9 \\ \hline
\textbf{\begin{tabular}[c]{@{}l@{}}Star Magnitude\\ (Brightest)\end{tabular}} & \multicolumn{1}{l|}{\cellcolor[HTML]{EFEFEF}\begin{tabular}[c]{@{}l@{}}2.2\end{tabular}} & \multicolumn{1}{l|}{\cellcolor[HTML]{EFEFEF}\begin{tabular}[c]{@{}l@{}}3.1\end{tabular}} & \multicolumn{1}{l|}{\cellcolor[HTML]{EFEFEF}\begin{tabular}[c]{@{}l@{}}3.8\end{tabular}} & \multicolumn{1}{l|}{\cellcolor[HTML]{EFEFEF}\begin{tabular}[c]{@{}l@{}}3.6\end{tabular}} & \multicolumn{1}{l|}{\cellcolor[HTML]{EFEFEF}\begin{tabular}[c]{@{}l@{}}3.6\end{tabular}} & \multicolumn{1}{l|}{\cellcolor[HTML]{EFEFEF}\begin{tabular}[c]{@{}l@{}}3.6\end{tabular}} & \multicolumn{1}{l|}{\cellcolor[HTML]{EFEFEF}\begin{tabular}[c]{@{}l@{}}0.0\end{tabular}} & \multicolumn{1}{l|}{\cellcolor[HTML]{EFEFEF}\begin{tabular}[c]{@{}l@{}}2.5\end{tabular}} & \multicolumn{1}{l|}{\cellcolor[HTML]{EFEFEF}\begin{tabular}[c]{@{}l@{}}3.1\end{tabular}} & \multicolumn{1}{l|}{\cellcolor[HTML]{EFEFEF}\begin{tabular}[c]{@{}l@{}}3.6\end{tabular}} & \multicolumn{1}{l|}{\cellcolor[HTML]{EFEFEF}\begin{tabular}[c]{@{}l@{}}3.6\end{tabular}} & \multicolumn{1}{l|}{\cellcolor[HTML]{EFEFEF}\begin{tabular}[c]{@{}l@{}}2.0\end{tabular}} & \cellcolor[HTML]{EFEFEF}\begin{tabular}[c]{@{}l@{}}3.8\end{tabular} \\ \hline
\textbf{} & \multicolumn{13}{c|}{\cellcolor[HTML]{9B9B9B}\textbf{Results (across/about) in arcseconds ( $\downarrow$ better)}} \\ \hline
\textbf{ICP~\cite{chin_star_2019}} & \multicolumn{1}{l|}{\begin{tabular}[c]{@{}l@{}}3.3E3   \\ 394.6\end{tabular}} & \multicolumn{1}{l|}{\begin{tabular}[c]{@{}l@{}}3.0E3   \\ 1.2E3\end{tabular}} & \multicolumn{1}{l|}{\begin{tabular}[c]{@{}l@{}}2.7E3   \\ 379.8\end{tabular}} & \multicolumn{1}{l|}{\begin{tabular}[c]{@{}l@{}}2.5E3   \\ 404.6\end{tabular}} & \multicolumn{1}{l|}{\begin{tabular}[c]{@{}l@{}}2.6E3   \\ 329.2\end{tabular}} & \multicolumn{1}{l|}{\begin{tabular}[c]{@{}l@{}}2.8E3   \\ 729.7\end{tabular}} & \multicolumn{1}{l|}{\begin{tabular}[c]{@{}l@{}}2.8E3\\ 680.8\end{tabular}} & \multicolumn{1}{l|}{\begin{tabular}[c]{@{}l@{}}1.6E3   \\ 2.2E3\end{tabular}} & \multicolumn{1}{l|}{\begin{tabular}[c]{@{}l@{}}1.6E3   \\ 2.1E3\end{tabular}} & \multicolumn{1}{l|}{\begin{tabular}[c]{@{}l@{}}591.3   \\ 3.1E3\end{tabular}} & \multicolumn{1}{l|}{\begin{tabular}[c]{@{}l@{}}713.8   \\ 3.4E3\end{tabular}} & \multicolumn{1}{l|}{\begin{tabular}[c]{@{}l@{}}2.3E3   \\ 6.2E3\end{tabular}} & \begin{tabular}[c]{@{}l@{}}1.2E3   \\ 2.5E3\end{tabular} \\ \hline
\textbf{Hough~\cite{bagchi_event-based_2020}} & \multicolumn{1}{l|}{\begin{tabular}[c]{@{}l@{}}8.6E3   \\ 8.7E4\end{tabular}} & \multicolumn{1}{l|}{\begin{tabular}[c]{@{}l@{}}5.0E3   \\ 4.8E4\end{tabular}} & \multicolumn{1}{l|}{\begin{tabular}[c]{@{}l@{}}2.6E4   \\ 5.2E4\end{tabular}} & \multicolumn{1}{l|}{\begin{tabular}[c]{@{}l@{}}3.5E4   \\ 3.4E4\end{tabular}} & \multicolumn{1}{l|}{\begin{tabular}[c]{@{}l@{}}3.1E4   \\ 2.5E4\end{tabular}} & \multicolumn{1}{l|}{\begin{tabular}[c]{@{}l@{}}1.8E4   \\ 5.9E4\end{tabular}} & \multicolumn{1}{l|}{\begin{tabular}[c]{@{}l@{}}1.1E3\\ 1.6E4\end{tabular}} & \multicolumn{1}{l|}{\begin{tabular}[c]{@{}l@{}}1.1E4   \\ 5.7E4\end{tabular}} & \multicolumn{1}{l|}{\begin{tabular}[c]{@{}l@{}}1.6E4   \\ 4.8E4\end{tabular}} & \multicolumn{1}{l|}{\begin{tabular}[c]{@{}l@{}}1.5E4   \\ 6.3E4\end{tabular}} & \multicolumn{1}{l|}{\begin{tabular}[c]{@{}l@{}}8.2E3\\ 2.9E4\end{tabular}} & \multicolumn{1}{l|}{\begin{tabular}[c]{@{}l@{}}1.1E4   \\ 6.5E4\end{tabular}} & \begin{tabular}[c]{@{}l@{}}7.6E4   \\ 7.3E4\end{tabular} \\ \hline
\textbf{2D-KF~\cite{Ng2022, latif_high_2023}} & \multicolumn{1}{l|}{\cellcolor[HTML]{FFCCC9}\begin{tabular}[c]{@{}l@{}}228.9   \\ 1.3E3\end{tabular}} & \multicolumn{1}{l|}{\cellcolor[HTML]{FFCCC9}\begin{tabular}[c]{@{}l@{}}236.4   \\ 1.0E3\end{tabular}} & \multicolumn{1}{l|}{\cellcolor[HTML]{FFCCC9}\begin{tabular}[c]{@{}l@{}}92.6   \\ 1.2E3\end{tabular}} & \multicolumn{1}{l|}{\cellcolor[HTML]{FFCCC9}\begin{tabular}[c]{@{}l@{}}81.9   \\ 1.0E3\end{tabular}} & \multicolumn{1}{l|}{\cellcolor[HTML]{FFCCC9}\begin{tabular}[c]{@{}l@{}}62.6   \\ 813.6\end{tabular}} & \multicolumn{1}{l|}{\cellcolor[HTML]{FFCCC9}\begin{tabular}[c]{@{}l@{}}67.9   \\ 1.3E3\end{tabular}} & \multicolumn{1}{l|}{\cellcolor[HTML]{FFCCC9}\begin{tabular}[c]{@{}l@{}}171.4\\ 2.1E3\end{tabular}} & \multicolumn{1}{l|}{\cellcolor[HTML]{FFCCC9}{\color[HTML]{333333} \begin{tabular}[c]{@{}l@{}}337.4   \\ 7.3E3\end{tabular}}} & \multicolumn{1}{l|}{\cellcolor[HTML]{FFCCC9}\begin{tabular}[c]{@{}l@{}}294.6   \\ 6.6E3\end{tabular}} & \multicolumn{1}{l|}{\cellcolor[HTML]{FFCCC9}\begin{tabular}[c]{@{}l@{}}846.0   \\ 1.6E4\end{tabular}} & \multicolumn{1}{l|}{\cellcolor[HTML]{FFCCC9}\begin{tabular}[c]{@{}l@{}}607.2   \\ 1.2E4\end{tabular}} & \multicolumn{1}{l|}{\cellcolor[HTML]{FFCCC9}\begin{tabular}[c]{@{}l@{}}485.1   \\ 9.2E3\end{tabular}} & \cellcolor[HTML]{FFCCC9}\begin{tabular}[c]{@{}l@{}}221.8   \\ 7.3E3\end{tabular} \\ \hline
\textbf{\begin{tabular}[c]{@{}l@{}}EBS-EKF \\ w/o offset\end{tabular}} & \multicolumn{1}{l|}{\cellcolor[HTML]{FFFC9E}\begin{tabular}[c]{@{}l@{}}27.0   \\ 83.1\end{tabular}} & \multicolumn{1}{l|}{\cellcolor[HTML]{FFFC9E}\begin{tabular}[c]{@{}l@{}}31.9   \\ 109.3\end{tabular}} & \multicolumn{1}{l|}{\cellcolor[HTML]{FFFC9E}\begin{tabular}[c]{@{}l@{}}25.9   \\ 84.5\end{tabular}} & \multicolumn{1}{l|}{\cellcolor[HTML]{FFFC9E}\begin{tabular}[c]{@{}l@{}}24.2   \\ 88.6\end{tabular}} & \multicolumn{1}{l|}{\cellcolor[HTML]{FFFC9E}\begin{tabular}[c]{@{}l@{}}26.0   \\ 80.8\end{tabular}} & \multicolumn{1}{l|}{\cellcolor[HTML]{FFFC9E}\begin{tabular}[c]{@{}l@{}}22.1   \\ 86.1\end{tabular}} & \multicolumn{1}{l|}{\cellcolor[HTML]{FFFC9E}\begin{tabular}[c]{@{}l@{}}172.1\\ 322.0\end{tabular}} & \multicolumn{1}{l|}{\cellcolor[HTML]{FFFC9E}\begin{tabular}[c]{@{}l@{}}57.3   \\ 64.8\end{tabular}} & \multicolumn{1}{l|}{\cellcolor[HTML]{9AFF99}\textbf{\begin{tabular}[c]{@{}l@{}}71.1   \\ 70.2\end{tabular}}} & \multicolumn{1}{l|}{\cellcolor[HTML]{9AFF99}\textbf{\begin{tabular}[c]{@{}l@{}}21.6   \\ 66.2\end{tabular}}} & \multicolumn{1}{l|}{\cellcolor[HTML]{FFFC9E}\begin{tabular}[c]{@{}l@{}}17.0   \\ 84.7\end{tabular}} & \multicolumn{1}{l|}{\cellcolor[HTML]{FFFC9E}\begin{tabular}[c]{@{}l@{}}50.3   \\ 79.1\end{tabular}} & \cellcolor[HTML]{FFFC9E}\begin{tabular}[c]{@{}l@{}}26.7   \\ 84.5\end{tabular} \\ \hline
\textbf{\begin{tabular}[c]{@{}l@{}}EBS-EKF \\ w/ offset\end{tabular}} & \multicolumn{1}{l|}{\cellcolor[HTML]{9AFF99}{\color[HTML]{000000} \textbf{\begin{tabular}[c]{@{}l@{}}25.8   \\ 60.3\end{tabular}}}} & \multicolumn{1}{l|}{\cellcolor[HTML]{9AFF99}\textbf{\begin{tabular}[c]{@{}l@{}}31.7   \\ 87.9\end{tabular}}} & \multicolumn{1}{l|}{\cellcolor[HTML]{9AFF99}\textbf{\begin{tabular}[c]{@{}l@{}}26.1   \\ 75.3\end{tabular}}} & \multicolumn{1}{l|}{\cellcolor[HTML]{9AFF99}\textbf{\begin{tabular}[c]{@{}l@{}}25.2   \\ 69.1\end{tabular}}} & \multicolumn{1}{l|}{\cellcolor[HTML]{9AFF99}\textbf{\begin{tabular}[c]{@{}l@{}}27.0   \\ 77.9\end{tabular}}} & \multicolumn{1}{l|}{\cellcolor[HTML]{9AFF99}\textbf{\begin{tabular}[c]{@{}l@{}}22.1   \\ 78.1\end{tabular}}} & \multicolumn{1}{l|}{\cellcolor[HTML]{9AFF99}\textbf{\begin{tabular}[c]{@{}l@{}}170.1\\ 139.0\end{tabular}}} & \multicolumn{1}{l|}{\cellcolor[HTML]{9AFF99}\textbf{\begin{tabular}[c]{@{}l@{}}57.4   \\ 52.8\end{tabular}}} & \multicolumn{1}{l|}{\cellcolor[HTML]{FFFC9E}\begin{tabular}[c]{@{}l@{}}70.9   \\ 74.1\end{tabular}} & \multicolumn{1}{l|}{\cellcolor[HTML]{FFFC9E}\begin{tabular}[c]{@{}l@{}}24.7   \\ 70.5\end{tabular}} & \multicolumn{1}{l|}{\cellcolor[HTML]{9AFF99}\textbf{\begin{tabular}[c]{@{}l@{}}18.6   \\ 82.9\end{tabular}}} & \multicolumn{1}{l|}{\cellcolor[HTML]{9AFF99}\textbf{\begin{tabular}[c]{@{}l@{}}49.1   \\ 64.5\end{tabular}}} & \cellcolor[HTML]{9AFF99}\textbf{\begin{tabular}[c]{@{}l@{}}28.6   \\ 75.9\end{tabular}} \\ \hline
\end{tabular}
\caption{Metadata and tracking results for our real world dataset. Metrics are reported as (across, about) mean squared difference in arcseconds. Colors rank methods with the first (green), second (yellow), or third (red) lowest across and about means.}
\label{table:main-results}
\vspace{-10pt}
\end{table*}

%% file: new_new_secs/6_conclusions.tex
\section{Discussion and Conclusion}
\noindent \textbf{Limitations:} Obtaining precise ground truth for star tracking presents challenges, as highlighted in prior star tracker studies \cite{gudmundson_ground_2019}. Some approaches achieve this by calibrating a motorized pan/tilt mount to reference stars and using the mount’s position readout as ground truth \cite{brady_ground_2004}, while others rely on purely simulated data for benchmarking \cite{chin_star_2019}. To assess performance in real-world conditions, we adopt the solution of benchmarking our system against a finely calibrated, high-accuracy commercial star tracker (the Rocket Lab APS tracker). Although our approach is bounded by the reference star tracker’s accuracy, it enables direct, meaningful comparisons between our system and a standard tracker on real data.

While our earth-based data contains some artifacts that wouldn’t be present in space—such as atmospheric diffraction variations near the horizon and shifts in background illumination --- their impact on our experimental results is expected to be small compared to the order of magnitude larger error observed in our benchmarking methods. Additionally, we verify our centroiding approach separately using synthetic (LCD monitor) stars with idealized ground truth to ensure this component of our algorithm remains unaffected by APS tracker inaccuracy or other external factors.

\noindent \textbf{Conclusion:} This work presents an event-based star tracking method that both quantitatively improves upon the accuracy of previous EBS approaches and extends the operating envelope relative to the motion tolerance and update rate of APS-based star trackers. The accuracy improvements result from our novel centroiding approach based on EBS pixel modeling in low light and our EBS-EKF algorithm's ability to estimate attitude in three dimensions. Both the sparsity of the event data and the sensor's asynchronous output allow us to achieve significantly higher update rates than our APS reference tracker, all while achieving comparable accuracy. %In some cases, we outperform the APS tracker, like shown in Figure~\ref{fig:teaser} where we track at higher angular velocities. 

While previous studies have highlighted the potential advantages of event cameras over APS trackers in proof-of-concept scenarios, we demonstrate these benefits using real data. We capture the first dataset of real events of real star fields with a commercially-available event sensor, and provide synchronized attitude estimates from a conventional star tracker as approximate ground truth. This data is released as a public benchmark to allow other researchers to produce comparable results in this growing area of research.

%% file: new_new_secs/4_experiments.tex
\section{Experimental Setup}
\label{sec:experiments}

In order to quantify the accuracy of our attitude estimates, and to characterize the expanded range of operating conditions where our event-based star tracker can operate, we created a hardware setup to capture real event data of real stars. Event streams from a Prophesee EVK4-HD sensor were synchronized with the outputs of an off-the-shelf APS-based star tracker as their fields of view were moved across the night sky.

\noindent \textbf{APS Hardware: Rocket Labs Star Tracker:} 
We utilize a RocketLabs ST-16RT2 star tracker as our reference attitude-determination solution. 
The ST-16RT2 consists of a 2592 x 1944 pixel CMOS active-pixel sensor, a custom 16mm f/1.6 lens, a 12 cm rigid sun-keepout baffle, and an onboard processor capable of providing attitude solutions at 2Hz over an RS-485 link which we access via a USB to RS-485 converter. 
The manufacturer states the tracker can achieve accuracies of 5 arcseconds RMS cross-boresight and 55 arcseconds RMS around-boresight, with a maximum slew rate of 3\textdegree/second. 
The system is set up for a field of view of 15\textdegree x 20\textdegree.

\indent \textbf{DVS Hardware: Prophesee EVK4-HD:} Our event camera is one of the latest from Prophesee, the EVK4-HD.
The EVK4-HD consists of a 1280x720 event-pixel array (pixels are 4.86um in size) and related readout circuitry which is accessible over USB. 
The system is set up with a 35mm lens focused at infinity with a field of view of 5.7\textdegree x 10.2\textdegree. 
This leads to an effective instantaneous field of view for each pixel of approximately 30 arcseconds. \connor{Perhaps of note is the high fill factor which helps with low-light}

\begin{figure}
    \centering
    \includegraphics[width=\linewidth]{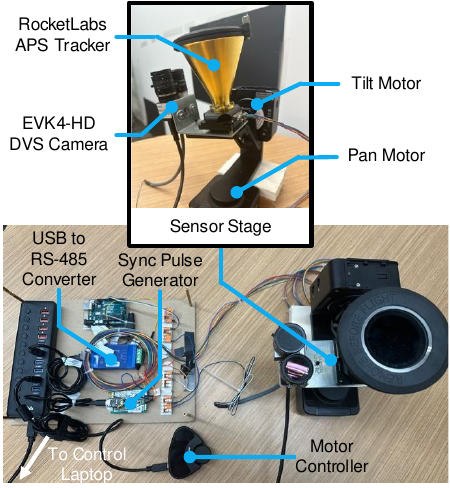}
    \caption{\textbf{Our data collection system with key components labelled.}
    }
    \label{fig:experimental:data_collect_rig}
\end{figure}

We collect earth-based data of sensor rotation to validate our estimation techniques. 
Our complete collection system consists of a Prophesee EVK4-HD and a commercially available RocketLabs star tracker mounted rigidly to a computer-controlled pan/tilt mechanism (Figure \ref{fig:experimental:data_collect_rig}). 
The system is controlled via a Windows laptop. 
After a short static calibration period, the pan/tilt mechanism is used to rotate both sensors at various speeds to simulate satellite rotation.

\begin{figure}
    \centering
    \includegraphics[width=\linewidth]{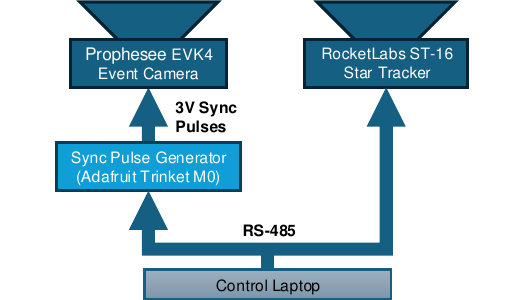}
    \caption{\textbf{Block diagram of our event camera to commercial star tracker synchronization mechanism.} 
    The control laptop generates requests for an attitude solution from the commercial star tracker. 
    The sync pulse generator listens for these requests, and fires off a synchronization pulse to the event which records the pulse alongside a timestamp in the event timescale.}
    \label{fig:experimental:sync_generator}
\end{figure}

\subsection{DVS/EVT Synchronization}

We synchronize our event camera with the Rocket Labs star tracker to align our orientation estimates temporally. 
To do so, we set up a small microcontroller (an Adafruit Trinket M0) which listens to the RS-485 serial line used to communicate with the RocketLabs tracker. 
When a solution is requested by our control computer via this serial line, the microcontroller sends a square wave pulse to the EVK4 which records this pulse in the event stream with the appropriate timestamp.
The RocketLabs solution is timestamped relative to the time of the solution request which is recorded in the event stream. 
This combination of recorded data allows us to identify the precise time when the RocketLabs-estimated attitude is valid in the event data timeline. 
Additionally, the same square wave pulse from the microcontroller is used to trigger an image acquisition by the Basler high speed camera.
This generates a sequence of framing images also aligned to the event camera timescale. 

\subsection{Relative Attitude Calibration}

To compute the differential between the Rocket Labs APS tracker and the DVS camera, we need to determine the relative rotation between the cameras. Assuming the cameras share a rigid body (i.e., mounted to the same tripod), the relative rotation remains constant regardless of the cameras' orientations. Therefore, we evaluate performance based on the deviation of the DVS tracker from this relative rotation. While stereo calibration using pixel values would be ideal, the Rocket Labs tracker provides quaternion solutions that are precisely calibrated to its imaging plane and undergo proprietary signal processing corrections inaccessible to the user. Consequently, we must rely directly on the quaternion solutions from the Rocket Labs tracker.

To obtain the relative rotation, we (1) perform velocity sweeps of the night sky with both cameras, as shown in the velocity sweep experiments in Figure~\ref{fig:results
}. We then run our tracker over these sweeps and identify the rotation that minimizes the error between the Rocket Labs solutions and our tracker’s solutions. Mathematically, we determine the relative rotation $q_r$ that minimizes the rotation error between the two trackers:

\begin{align} \min_{\mathbf{q}_r}\sum_i||\mathbf{q}_{\text{RL}}^i \cdot \mathbf{q}_r - \mathbf{q}_{\text{DVS}}^i|| \end{align}

Here, the minus sign represents the boxminus operator for subtracting rotations, as described in Hertzberg et al.~\cite{hertzberg_integrating_2011}. Crucially, the choice of tracker has minimal impact on the relative rotation, as any error is averaged out over the velocity sweeps in both directions. We measured the consistency of our relative rotation across three velocity sweeps to be within 20 arcseconds. We use the same relative rotation on all tracks so there is no data leakage. 

\subsubsection{Offset Curve Calibration}

Additionally, we use the velocity sweeps to derive the offset curves needed to correct for the offsets described in Section~\ref{sec:event-signal-model}. Specifically, we compute the event profile for stars of varying magnitudes, as illustrated in Figure~\ref{fig:methods:event_profile} and plot the offset of the event mean against the star magnitude as in Figure~\ref{fig:methods:event_profile}d. For our experiments, we generate the offset calibration curve using data from one track and apply this curve across all tracks.

\subsection{APS and DVS Synchronized Tracking Setup}

\textbf{Methods:} We implemented our method along with the DVS algorithms described in the Related Works section~\ref{sec:dvs-trackers}. Due to the lack of open-source code provided with the original papers, we re-implemented all methods from scratch. To optimize each method's performance, we conducted a parameter sweep on the hyperparameters. For the ICP~\cite{chin_star_2019} and Hough~\cite{bagchi_event-based_2020} methods, we provide absolute measurements every 10 seconds of the track as these methods quickly drift off track with them. We implement the Hough method with a resolution of 30 ms, which we find works best for our data. For the ICP method, we estimate frame to frame rotations using a binning time of 30 ms, which was determined best using a hyperparameter sweep. The 2D-KF and our EVT-EKF methods only require an absolute measurement at the beginning of the track. 

\begin{figure*}
    \centering
    \includegraphics[width=\linewidth]{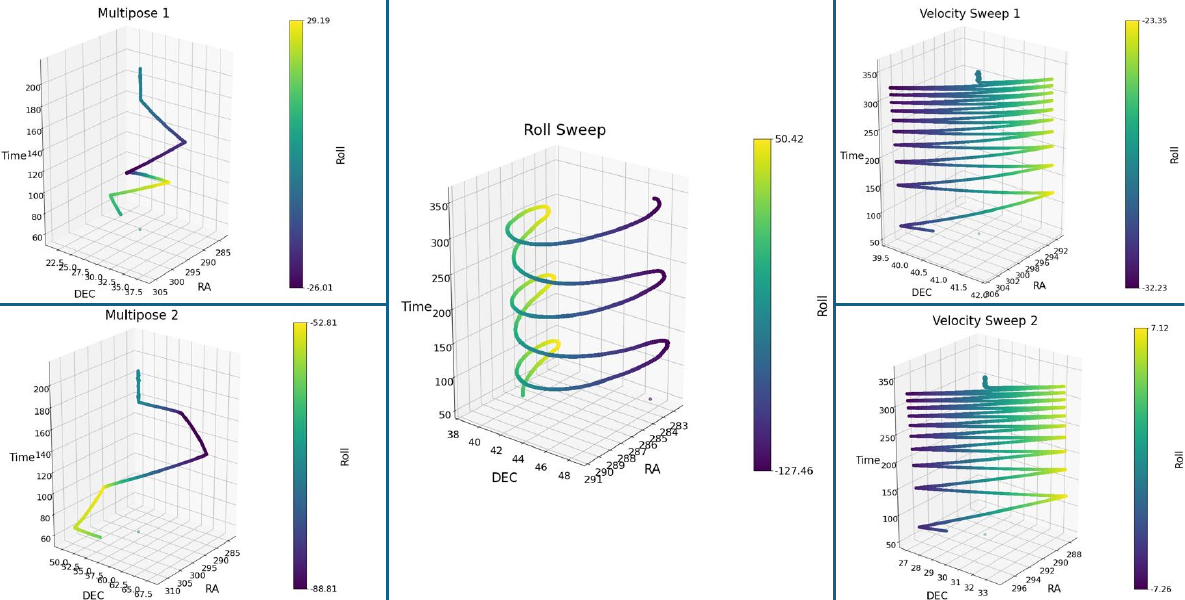}
    \caption{\textbf{Visualizations of the motion profiles estimated by our tracker.}
    Right Ascension and Declination are shown along the bottom, and time flows vertically from top to bottom. 
    Roll is indicated by the color of the line.}
    \label{fig:experimental:track_profiles}
\end{figure*}

\noindent \textbf{Tracks:} To benchmark the methods, we captured 5-6 tracks, each lasting between 2 to 5 minutes. The tracks were recorded using various camera motions and velocities to evaluate the methods under different conditions. Specifically, the angular velocities for the velocity sweeps and multipose tracks ranged from 0.1 to 3 deg/sec in the yaw and pitch directions (i.e., $x$ and $y$). For the roll sweep, the angular velocities for yaw and pitch were kept constant, while the roll angular velocity was increased up to 20 deg/sec. Lastly, we conducted a high-velocity sweep, increasing yaw and pitch velocities up to $7.5$ deg/sec, representing our most challenging track due to the significant linear velocities imposed on the image plane.
Visualizations of these tracks are shown in Figure~\ref{fig:experimental:track_profiles}, and key characteristics are shown in Table~\ref{table:experimental:track_char}.

\begin{table*}[]
\centering
\begin{tabular}{l|l|l|l}
Experiment Name  & Velocity Range (\textdegree / s) & RA/Dec/Roll Range (\textdegree) & Time (s)                 \\ \hline
Velocity Sweep 1 & 0.28-2.8                        & 13.92/2.28/8.88                & \multicolumn{1}{l|}{290} \\
Velocity Sweep 2 & 0.28-2.8                        & 9.75/6.64/14.38                & \multicolumn{1}{l|}{290} \\
Multipose 1      & 0.84-0.84                       & 21.72/17.21/55.20              & \multicolumn{1}{l|}{151} \\
Multipose 2      & 0.84-0.84                       & 28.09/18.55/36.00              & \multicolumn{1}{l|}{150} \\
Roll Sweep       & 14-20                           & 7.99/10.09/177.88              & \multicolumn{1}{l|}{290}
\end{tabular}
\caption{Summary of key characteristics of each experimental track.
Velocity range is determined by the minimum and maximum commanded velocities sent to the pan/tilt system.
RA/Dec/Roll range is computed from our EVT-EKF estimates.}
\label{table:experimental:track_char}
\end{table*}

\noindent \textbf{Metrics:} We reported the mean squared difference between the Rocket Labs tracker and the DVS tracker for both pointing accuracy (across) and roll accuracy (about), as is standard in star tracking literature~\cite{liebe_accuracy_2002}. The Rocket Labs tracker is rated for 5 arcseconds across and 50 arcseconds about at slew rates up to 3 deg/sec in yaw and pitch. It is important to note that these quantitative metrics do not represent errors relative to ground truth, as the Rocket Labs tracker itself has inherent error margins. Nevertheless, the observed differences provide a solid understanding of the methods' performance within the 5/50 arcsecond accuracy range of the Rocket Labs tracker.

%% file: new_new_secs/X_suppl.tex
\clearpage
\setcounter{page}{1}
\maketitlesupplementary

\section{Overview}

We include a video titled \textit{Short video overview.mp4} along with this document, which provides a narrated overview of our paper.
Section~\ref{supp:ekf} derives our extended Kalman filter (EKF) for event-based star tracking. Section~\ref{supp:additional-tracking-details} provides additional details on our tracking algorithm. Section~\ref{supp:additional-centroiding} provides implementation details and a discussion on our laboratory setup used to perform centroiding experiments. Section~\ref{supp:additional-nightsky} provides more details on our synchronized EVK4-HD event camera and APS star tracker, including their hardware configurations, synchronization scheme, and procedures for obtaining the brightness-dependent offset curve. Section~\ref{supp:night-sky-dataset} provides additional details on our night sky dataset. Section~\ref{supp:additional-tracking-results} provides plots for the tracks summarized in Table~\ref{table:main-results} of the main text.

\section{State estimation with an Extended Kalman Filter (EKF)}
\label{supp:ekf}

This section provides more details on estimating our state using an EKF. Restated from the main text, our measurements are the set of $N$, positive events $\mathcal{E} = \{\mathbf{x}, t\}_{i=0:N}$, where we denote each event as $\event \in \mathcal{E}$. 

\subsection{Bayesian State Estimation with a Kalman Filter}

Under a Markov assumption, the probability distribution over all states and measurements is, 
\begin{equation}
p(\state_0,...,\state_N; e_0,...,e_N) = p(\state_0)\prod_{i=0}^{N}\mathcal{L}(e_i | \state_i)p(\state_i | \state_{i - 1}).
\end{equation}
The Kalman filter uses predict and update phases to recursively estimate the posterior distribution of the states conditioned on measurements up to the current timestep. As shown in ~\cite{barker1995bayesian}, the predicted state is the distribution associated with the previous state, over all possible previous states,
\begin{equation}
p(\state_i | \mathcal{E}_{i - 1}) = \int p(\state_i | \state_{i - 1})p(\state_{i - 1} | \mathcal{E}_{i -1}) d\state_{i - 1}
\end{equation}
where the measurements up to the current timestep $t$ are $\mathcal{E}_t = \{e_1, ..., e_t\}$. The update step is the product of the measurement likelihood and the predicted state,
\begin{equation}
p(\state_i | \mathcal{E}_{i - 1}) = \frac{\int p(e_i | \state_i)p(\state_i | \mathcal{E}_{i - 1})d \state_i}{p(\state_i | \mathcal{E}_{i - 1})}
\end{equation}
We refer interested readers to Barker et al.,~\cite{barker1995bayesian} for more details on these distributions and how they are estimated by the EKF equations that we define in the next section.

\subsection{EKF Predict and Update Matrix Equations}

\subsubsection{Measurement Geometry}

We aim to use EBS measurements of stars to estimate the camera state $\state = [\quat, \mathbf{\omega}]$, where $\quat \in SO(3)$ is a quaternion encoding a valid 3D rotation, and $\mathbf{\omega} \in \real^3$ denotes the 3D angular velocity. For our work, we consider the catalog of stars $\mathbf{S} \in \real^{N \times 3}$ with apparent magnitude $\leq 7$. Each of the $N$ stars is a 3D point normalized to the unit sphere with negligible parallax effects~\cite{forbes2015fundamentals}. A single star $\mathbf{s}_i \in \mathbf{S}$ with 3D coordinates $(X, Y, Z)$ is rotated into the camera's frame using a world-to-camera quaternion $\attitude_{\text{wc}}$\footnote{We typically omit the wc subscript for brevity.}, 

\begin{align}
\hat{\mathbf{s}}_i = \attitude_{\text{wc}}(\mathbf{s}_i), \label{supp:measurement-fn-1}
\end{align}
and projected onto the 2D imaging plane using the pinhole camera model
\begin{align}
    x = f\frac{X}{Z}, \;\;\; y = f\frac{Y}{Z}, \label{supp:measurement-fn-2}
\end{align}
where $f$ is the focal length.

\subsubsection{EKF Predict and Update Equations}

Our state is a compound manifold defined by $SO(3) \times \real^3$ (i.e., the camera's rotation and angular velocity). To estimate our state with an EKF, we employ the boxplus operator $\boxplus$~\cite{hertzberg_integrating_2011}, which defines the addition of a 6D residual vector $[\delta_{\theta_x}, \delta_{\theta_y}, \delta_{\theta_z}, \delta_{\omega_x}, \delta_{\omega_y}, \delta_{\omega_z}]$ to our state manifold. Formally, the operator is defined as, 
\begin{equation}
\boxplus : \state \times \real^6 \rightarrow \state,
\end{equation}
and provides a mechanism for adding a change in rotation and velocity to our state quaternion and velocity.

We use a constant velocity prior to have the state attitude respond proportionally to the velocity and time,
\begin{align}
    \quat^{t+1} = \quat^t \boxplus \Delta t \omega = \exp  (\Delta t \omega )\cdot\quat^t \label{eq:quat-transition}.
\end{align}
Here, the boxplus operator uses the exponential map to project a change in angle to the quaternion's Lie group, where it is then multiplied with the quaternion to update the rotation~\cite{hertzberg_integrating_2011}. 
% \connor{Note, the constant velocity prior is reasonable under short time frames in which event-based tracking is performed in.}
With a constant velocity prior defined, the estimation of the state mean $\mathbf{u}$ is given by the standard EKF equations,
\begin{align}
    \bar{\mathbf{u}} = f(\mathbf{u}) \label{eq:process-ekf}, \\
    \mathbf{F} = \odv{f(\mathbf{u}^t)}{\mathbf{u}}_{\mathbf{u}^t}^{} \label{eq:process_jacobian-ekf}, \\
    \bar{\mathbf{P}} = \mathbf{F}\mathbf{P}\mathbf{F}^\top + \mathbf{Q}. \label{eq:covariance_predict-ekf}
\end{align}
Eq.~\ref{eq:process_jacobian-ekf} defines the Jacobian for the constant velocity model $f(\cdot)$ at time $t$. Eq.~\ref{eq:quat-transition} uses the model to predict the state mean $\bar{\mathbf{u}}$. Eq.~\ref{eq:covariance_predict-ekf} updates the state uncertainty $\mathbf{P}$ using the linearized model $\mathbf{F}$ and process uncertainty $\mathbf{Q}$. 

The EKF update phase is defined with the following equations:
\begin{align}
    \mathbf{H} = \odv{h(\bar{\ut})}{\mathbf{u}}_{\ut}^{} \label{eq:measurement_fn-ekf}, \\
    \mathbf{y} = \mathbf{z} - h(\bar{\mathbf{u}}), \label{eq:residual-ekf} \\
    \mathbf{K} = \bar{\mathbf{P}}\mathbf{H}^\top(\mathbf{H}\bar{\mathbf{P}}\mathbf{H}^\top + \mathbf{R})^{-1}, \label{eq:kalman_gain-ekf} \\
    \mathbf{u} = \bar{\mathbf{u}} \boxplus \mathbf{K}\mathbf{y}, \label{eq:update_mean-ekf} \\
    \mathbf{P} = (\mathbf{I} - \mathbf{K}\mathbf{H})\bar{\mathbf{P}}. \label{eq:update_covariance-ekf}
\end{align}
Eq.~\ref{eq:measurement_fn-ekf} defines a Jacobian for the measurement model $h(\cdot)$. Eq.~\ref{eq:residual-ekf} is the residual between a measurement $\mathbf{z}$ and the predicted state measurement $h(\bar{\mathbf{x}})$. Eq.~\ref{eq:kalman_gain-ekf} defines the Kalman gain, which is used in Eq.~\ref{eq:update_mean-ekf} to interpolate the prior and measurement via the boxplus operator. Finally, Eq.~\ref{eq:update_covariance-ekf} uses the Kalman gain to update the state covariance. 

\subsection{Derivation of F, Q, H}

\subsubsection{Forward model Jacobian:} The forward model jacobian is a matrix $\mathbf{F} \in \real^{6 \times 6}$ that linearizes Eq~\ref{eq:quat-transition}. Our Jacobian matrix is comprised of four partial derivatives,
\begin{align}
\mathbf{F} = \begin{bmatrix}
\odv{\quat^{t+1}}{\quat^t} & \odv{\quat^{t+1}}{\omega^t} \\
\odv{\omega^{t+1}}{\quat^t} & \odv{\omega^{t+1}}{\omega^t}.
\end{bmatrix}
\end{align}
Using identities defined by Bloesch et al.~\cite{bloesch_primer_2016}, we derive the first partial derivative w.r.t the attitude as 
\begin{align}
 & \odv{\quat^{t+1}}{\quat^t} = \odv{}{\quat^t}\left[\text{exp}(\Delta t \omega^{t}) \cdot \quat^{t} \right] = \mathbf{C}(\text{exp}(\Delta t \omega^{t})) \\ 
& = \mathbf{I} + \frac{\sin(||\Delta t \omega^{t}||)(\Delta t \omega^{t})^{\times}}{||\Delta t \omega^{t}||} \\ & \quad \quad \quad \quad \quad \quad \quad \quad + \frac{(1 - \cos(||\Delta t \omega^{t}||)(\Delta t \omega)^{{\times}^2}}{||\Delta t \omega^{t}||^2} \nonumber \\ & \approx \mathbf{I} + (\Delta t \omega^{t})^{\times}
\end{align}

% \begin{align}
% \odv{\quat^{t+1}}{\quat^t} &= \odv{}{\quat^t}\left[\text{exp}(\Delta t \omega^{t}) \cdot \quat^{t} \right] = \mathbf{C}(\text{exp}(\Delta t \omega^{t})) \\ 
% &= \mathbf{I} + \frac{\sin(||\Delta t \omega^{t}||)(\Delta t \omega^{t})^{\times}}{||\Delta t \omega^{t}||} \\
% & \; \; \; \; \; \; \;  + \frac{(1 - \cos(||\Delta t \omega^{t}||)(\Delta t \omega)^{{\times}^2}}{||\Delta t \omega^{t}||^2} \nonumber \\ & \approx \mathbf{I} + (\Delta t \omega^{t})^{\times}
% \end{align}

where $\mathbf{I} \in \real^{3 \times 3}$ is the identity matrix, $\mathbf{C}(\cdot)$ is Rodriguez' formula applied to an exponential mapped rotation, as defined by Bloesch et al.~\cite{bloesch_primer_2016}, and $\mathbf{v}^\times = (v_1, v_2, v_3)^{\times}$ denotes the $3\times3$ skew-symmetric matrix,
\begin{align}
\mathbf{v} = \begin{bmatrix}
0 & -v_3 & v_2 \\
v_3 & 0 & -v_1 \\
-v_2 & v_1 & 0
\end{bmatrix}.
\end{align}
We then derive the partial derivative with respect to velocity. Using the chain rule,
\begin{align}
\odv{\quat^{t+1}}{\omega^t} = \odv{}{\omega^t}\left[\text{exp}(\Delta t \omega^t) \cdot \quat^{t} \right] = \mathbf{\Gamma}(\Delta t \omega^t) \Delta t, 
\end{align}
where $\mathbf{\Gamma}$ is the Jacobian of the exponential map~\cite{bloesch_primer_2016} and expanded as
\begin{align}
\mathbf{\Gamma}(\Delta t \omega^t) = & \mathbf{I} +  \frac{(1 - \cos(||\Delta t \omega^t||))(\Delta t \omega^t)^{\times}}{||\Delta t \omega^t||^2} + \\ 
& \frac{(||\Delta t \omega^t|| - \sin(||\Delta t \omega^t||))(\Delta t \omega^t)^{\times^{2}}}{||\Delta t \omega^t||^3} \\
& \approx (\mathbf{I} + 1/2(\Delta t \omega^{t})^\times).
\end{align}
For the velocity transition function $\omega^{t+1}=\omega^{t}$, the derivative with respect to the attitude is simply $\odv{\omega^{t+1}}{\quat^t} = 0_{3 \times 3}$, and the derivative with respect to velocity is the identity matrix, $\odv{\omega^{t+1}}{\omega^t} = \mathbf{I}_{3 \times 3}$.
Filling in the four partial derivatives yields the $\mathbf{F}$ matrix,
\begin{small}
\[
\scalebox{0.8}{$
\mathbf{F} = \begin{bmatrix}
1 & -\Delta t \cdot \omega_3 & \Delta t \cdot \omega_2 & \Delta t & -\frac{(\Delta t)^2 \cdot \omega_3}{2} & \frac{(\Delta t)^2 \cdot \omega_2}{2} \\
\Delta t \cdot \omega_3 & 1 & -\Delta t \cdot \omega_1 & \frac{(\Delta t)^2 \cdot \omega_3}{2} & \Delta t & -\frac{(\Delta t)^2 \cdot \omega_1}{2} \\
-\Delta t \cdot \omega_2 & \Delta t \cdot \omega_1 & 1 & -\frac{(\Delta t)^2 \cdot \omega_2}{2} & \frac{(\Delta t)^2 \cdot \omega_1}{2} & \Delta t \\
0 & 0 & 0 & 1 & 0 & 0 \\
0 & 0 & 0 & 0 & 1 & 0 \\
0 & 0 & 0 & 0 & 0 & 1
\end{bmatrix}.
$}
\]
\end{small}

\subsubsection{Process noise:} The process noise matrix $\mathbf{Q} \in \real^{6 \times 6}$ adds our uncertainty in the constant velocity model to the Kalman filter covariance. The continuous white noise is applied to the velocity terms,
\begin{align}
\mathbf{Q}_c = 
\begin{bmatrix}
0 & 0 & 0 & 0 & 0 & 0 \\
0 & 0 & 0 & 0 & 0 & 0 \\
0 & 0 & 0 & 0 & 0 & 0 \\
0 & 0 & 0 & 1 & 0 & 0 \\
0 & 0 & 0 & 0 & 1 & 0 \\
0 & 0 & 0 & 0 & 0 & 1 \\
\end{bmatrix} \phi_s
\end{align} 
where $\phi_s$ is the power spectral density. In practice, the power spectral density is an engineering knob --- larger values result in smoother state predictions. The discretization of the noise through our process model~\cite{stacey2022analytical} is given by
\begin{align}
\mathbf{Q} = \int_0^{\Delta t} F(t) \mathbf{Q}_c F^{\top}(t) dt
\end{align}
We solve for $\mathbf{Q}$ using a symbolic solver, obtaining the matrix
\begin{small}
\[
\scalebox{0.8}{$
\mathbf{Q} \approx \begin{bmatrix}
0 & 0 & 0 & \frac{(\Delta t)^2}{2} & 0 & 0 \\
0 & 0 & 0 & 0 & \frac{(\Delta t)^2}{2} & 0 \\
0 & 0 & 0 & 0 & 0 & \frac{(\Delta t)^2}{2} \\
\frac{(\Delta t)^2}{2} & 0 & 0 & \Delta t & 0 & 0 \\
0 & \frac{(\Delta t)^2}{2} & 0 & 0 & \Delta t & 0 \\
0 & 0 & \frac{(\Delta t)^2}{2} & 0 & 0 & \Delta t
\end{bmatrix},
$}
\]
\end{small}
which is shown as an approximation because we have set small terms to zero. 

\subsubsection{Measurement Jacobian:} The event measurements relate to the attitude using the rotation and projection models of Eq.~\ref{supp:measurement-fn-1} and Eq.~\ref{supp:measurement-fn-2}. We define our measurement matrix $\mathbf{H} \in 2 \times 6$ for a single event as
\begin{align}
\mathbf{H} = \begin{bmatrix}
\odv{x}{\quat^t} & 0_{1 \times 3} \\
\odv{y}{\quat^t} & 0_{1 \times 3},
\end{bmatrix}
\end{align}
where $\odv{x}{\quat^t} \in \real^{1 \times 3}$, $\odv{y}{\quat^t} \in \real^{1 \times 3}$, noting that we do not incorporate velocity into our measurement model (i.e., $\odv{x}{\omega^t} = 0_{1 \times 3}$ and $\odv{y}{\omega^t} = 0_{1 \times 3}$).
Considering the event $x$ and $y$ coordinates, the partial derivatives w.r.t. attitude expands as
\begin{align}
    & \odv{x}{\quat^t} = \odv{}{\quat^t}\left[ f\frac{X}{Z} \right] = \odv{}{\quat^t} \left[ f\frac{\quat^t(\sworld)_x}{\quat^t(\sworld)_z} \right]
\end{align}
and
\begin{align}
    & \odv{y}{\quat^t} = \odv{}{\quat^t}\left[ f\frac{Y}{Z} \right] = \odv{}{\quat^t} \left[ f\frac{\quat^t(\sworld)_y}{\quat^t(\sworld)_z} \right],
\end{align}
respectively. In these equations, the quaternion $\quat(\cdot)$ rotates the star world coordinates to the camera, and the $(x, y, z)$ subscripts index that dimension of the rotated vector (e.g., $\quat^t(\sworld)_x = \quat^t(\sworld) \begin{bmatrix} 1 \\ 0 \\ 0 \end{bmatrix}$).  For completeness, the rotation of a vector $\mathbf{s} \in \real$ by a quaternion $\mathbf{q} = [q_0, \Breve{\mathbf{q}}] = [q_0, q_x, q_y, q_z]$ is defined as,

\begin{align}
\mathbf{q}(\mathbf{s}) = (2q_0^2 - 1)\mathbf{s} + 2q_0\Breve{\mathbf{q}}^{\times}\mathbf{s} + 2\Breve{\mathbf{q}}(\Breve{\mathbf{q}}^{\top}\mathbf{s}).
\end{align}

Using the product and chain rules, we compute the derivative of the event location $(x, y)$ with respect to attitude as
\begin{align}
     \odv{x}{\quat^t} = 
     f\left[-\odv{Z}{\quat^t}XZ^{-2} + \odv{X}{\quat^t} Z^{-1}\right]
\end{align}
and 
\begin{align}
     \odv{y}{\quat^t} = 
     f\left[-\odv{Z}{\quat^t}YZ^{-2} + \odv{Y}{\quat^t} Z^{-1}\right]
\end{align}
where the Jacobians of the rotations are given by 
\begin{align}
\odv{X}{\quat^t} = -(\quat(\mathbf{s}))^{\times}_x, \\
\odv{Y}{\quat^t} = -(\quat(\mathbf{s}))^{\times}_y, \\
\end{align}
and
\begin{align}
\odv{Z}{\quat^t} = -(\quat^t(\mathbf{s}))^{\times}_z.
\end{align}

\section{Tracking Algorithm Details}
\label{supp:additional-tracking-details}

We now provide more details for our tracking algorithm, which is summarized in the main text in Alg.~\ref{alg:tracking-alg}. We restate the algorithm here, 
\begin{algorithm}
\caption{EBS-EKF Star Tracking}\label{alg:cap}
\begin{algorithmic}
\State \textbf{Input: } Positive event stream $\mathcal{E} = \{\mathbf{x}, t\}_{i=0:N}$, 
\State \textbf{Output: } Camera State $\state$ (3D rotation and angular velocity)

\State \textbf{Initialize:} $\state_0 \gets \text{astrometry with binned positive events}$

%\State $k \gets 0$
\State $\text{r} \gets$ \text{max search radius}

\For{each event $e_i = (\mathbf{x}_i, t_i)$}

    \State $\hat{\state}_i \gets \text{EKF predict at $t_i$}$ 

    \State $\mathbf{x}_s \gets \text{closest projected catalog star to } \mathbf{x}_i$
    \If{distance($\mathbf{x}_i$, $\mathbf{x}_s$) $\leq$ r}
        \State $\bar{\mathbf{v}}_s \gets$ \text{star's normalized linear velocity via $\hat{\state}_i$}
        \State $m_s \gets$ \text{star's apparent magnitude}
        \State $\hat{\mathbf{x}}_i \gets \mathbf{x}_i + \bar{\mathbf{v}}_s \cdot z(m_s)$ \# apply offset correction
        \State $\state_i \gets \text{EKF update with $\hat{\mathbf{x}}_i$}$
    \EndIf
\EndFor
\end{algorithmic}
\label{alg:supp-tracking-alg}
\end{algorithm}
and provide a more detailed description of its operation. 

Referring to Alg.~\ref{alg:supp-tracking-alg}, we initialize the camera attitude by binning events with a small time window (e.g., 30 ms), identifying event clusters with DBSCAN clustering~\cite{ester_density-based_nodate}. Specifically, we accumulate positive events in batches of 60 milliseconds, and use \textit{scikit-learn}'s implementation of DBSCAN~\cite{scikit-learn} with parameters epsilon=2 and min\_samples=3. We pass the centroids of identified clusters to astrometry~\cite{lang_astrometry_2010}, which attempts to compute a camera attitude. We run astrometry with log odds of 14.

After obtaining an initial attitude with astrometry, we begin processing events using our EKF algorithm. First, we use the event's timestamp to update the EKF prediction using Eqs.~\ref{eq:process-ekf},~\ref{eq:process_jacobian-ekf},~\ref{eq:covariance_predict-ekf}. We then check if the event is within a set pixel radius of a catalog star, given the current attitude. This requires projecting catalog stars onto the imaging plane using the camera state. To do so, we use Eq.~\ref{supp:measurement-fn-1} to rotate the catalog with the current state, and Eq.~\ref{supp:measurement-fn-2} to project catalog stars into pixel space. If an event has a nearby catalog star, we use the star's magnitude and linear velocity to apply our intensity-dependent offset correction. We compute the linear velocity using the image Jacobian,

\begin{align}
\begin{bmatrix} v_x \\ v_y \end{bmatrix} = \begin{bmatrix} -xy/f & f + x^2/f & y \\ -f -y^2/f & xy/f & -x \end{bmatrix} \begin{bmatrix} \omega_x \\ \omega_y \\ \omega_z \end{bmatrix}
\label{eq:methods:star_vel}
\end{align}
where $\omega$ is known, since it is part of our EKF state. After applying the offset correction to the event, we use it as a measurement to update the EKF using Eq.~\ref{eq:update_mean-ekf} and~\ref{eq:update_covariance-ekf}.

\section{Monitor Centroiding Details}
\label{supp:additional-centroiding}

\noindent \textbf{Monitor Details:}
The 500 Hz monitor used is the ASUS ROG Swift Pro PG248QP. An important quality of this monitor is that there is minimal pixel value overshoot in-between frames (explained in \cite{rtings_monitor}). This lends frame transitions that are similar to real analog scenes. We did explore using OLED monitors but noticed that the event camera picked up on its refresh rate pattern.

\begin{figure}
  \centering
\includegraphics[width=.99\linewidth, page=1]{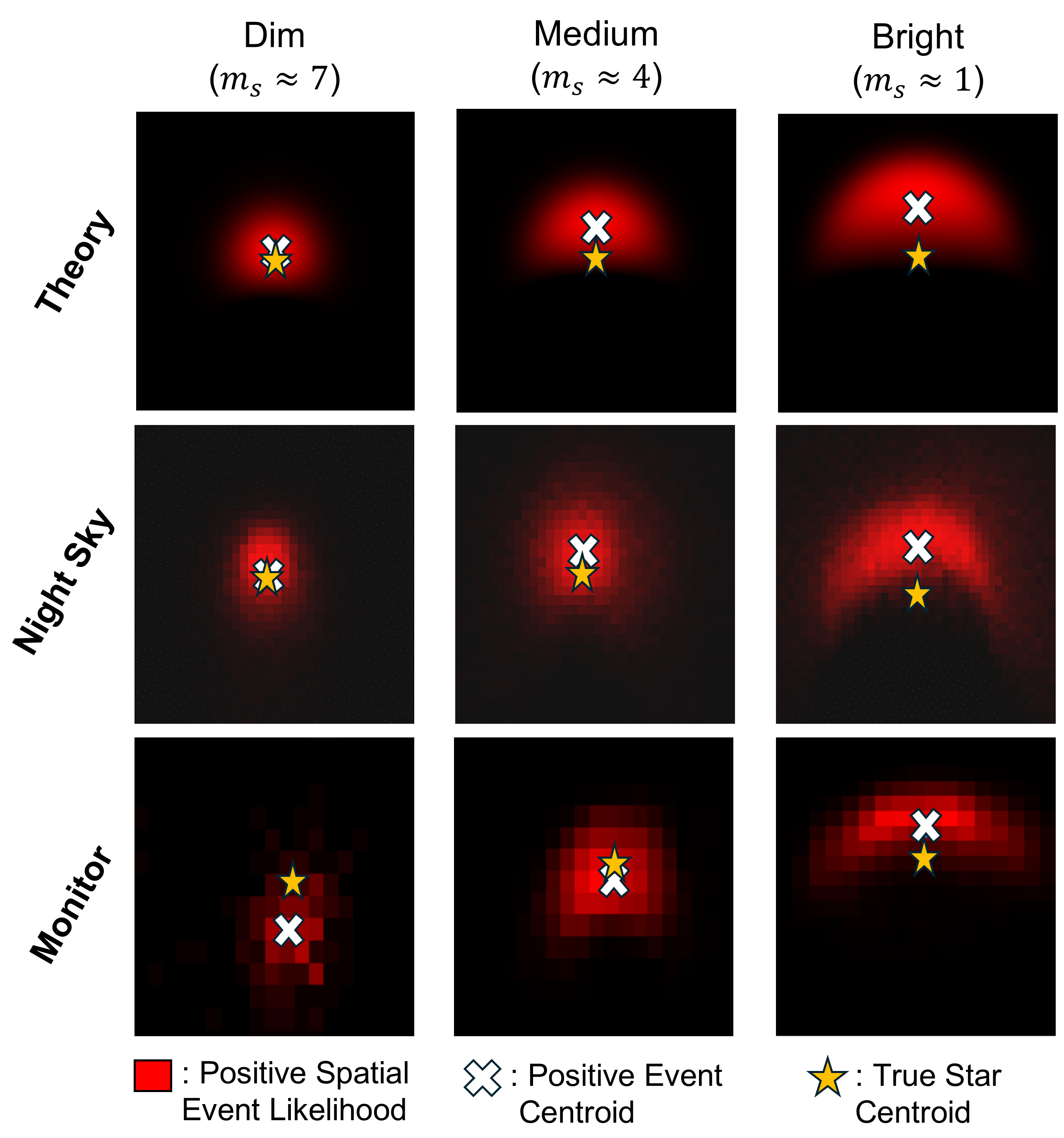}
   \caption{Example spatial event likelihoods from theory, night sky, and monitor measurements. This is similar to results shown in Fig \ref{fig:methods:event_profile} of the main paper. Note that the positive event centroid (white ``x'') lags behind the true star centroid more in the monitor experiments (3rd row) due to minor discrepancies in monitor/camera synchronization. However, the trend of the positive event centroid offset increasing with star magnitude is consistent across all modalities.}
   \label{supp:fig:monitor_stars}
\end{figure}

\noindent \textbf{Simulating Stars on Monitor:}
To simulate a moving star on the monitor, we display a Gaussian distribution as shown in Eq.~\ref{eq:method:gauss_star} in the paper. We use a 2 pixel standard deviation $\sigma_s$ (2.5 pixels on the camera's sensor) with a constant, unchanging speed across the monitor ($v_x=35$ sensor pixels per second, $v_y=0$). The pixel intensity values are also gamma-corrected so that it appears truly Gaussian to the camera sensor. To simulate different intensities stars, we use a combination of lowering the peak pixel intensity (e.g. 255 to 128 to 64) and by placing neutral-density (ND) filters in front of the camera which reduce the gathered light. As a result, each subsequently displayed star in the experiment is half as bright as the previously displayed star. We then have to calibrate the relative magnitude $m_s$ of just one of the displayed stars (found by matching the observed event rate to what we found in our night sky dataset) and the rest can be easily calculated. Examples of the monitor star event profiles are depicted in Fig.~\ref{supp:fig:monitor_stars}.

\noindent \textbf{Pixel Synchronization:}
The event camera was placed about 8 feet away from the monitor such that one sensor pixel corresponded to approximately one monitor pixel. To calculate the homography matrix between the sensor and the monitor, we flashed a series of eleven dots on the screen and binned the positive events with 10ms time bins. Knowing the position of the dots on the monitor and calculating the position of the dots on the sensor with the binned frames, we could recover a transformation matrix which relates the two coordinate frames. For this matrix, our mean reprojection error was approximately 0.5 pixels.

\begin{figure}
  \centering
\includegraphics[width=.99\linewidth, page=1]{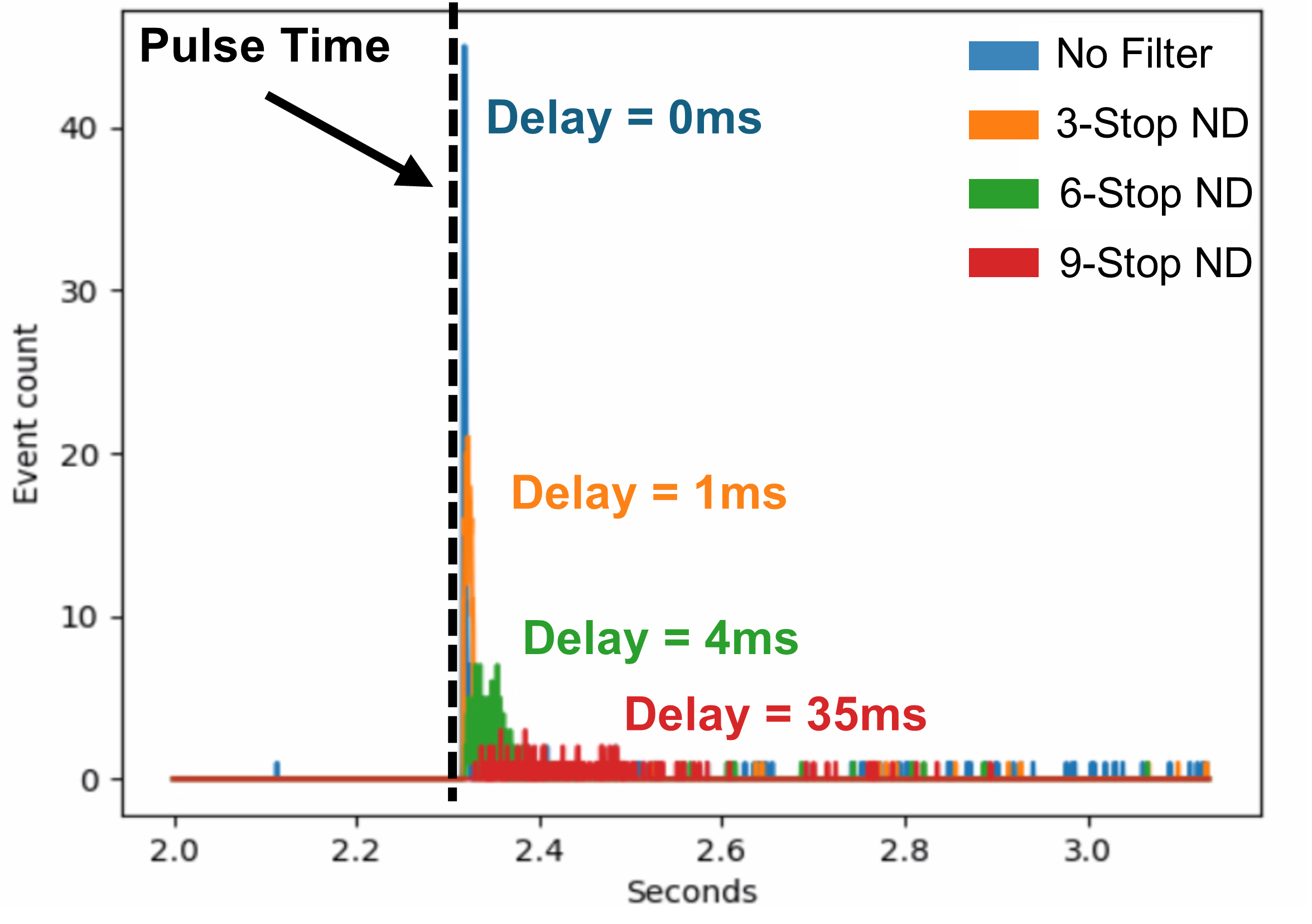}
   \caption{Synchronization timings of monitor experiments. Each neutral-density (ND) filter stop decreases light by 87.5\%. }
   \label{supp:fig:supp_synch}
\end{figure}

\noindent \textbf{Time Synchronization:}
In order to compare events to ground truth star locations in the monitor, we need to align the monitor and event time scales. To do so, we flash a set of pixels on the monitor at the start of an experiment --- the events generated from these pixels correspond with the start of the experiment. However, once we place ND filters over the camera, the events from these trigger pixels become delayed due to the low-pass filter effects discussed in the main paper and in~\cite{lichtsteiner_128_2008}. If unaccounted for, this would cause a systematic bias in our low light experiments. Thus, we perform a calibration procedure where we place ND filters over sections of the monitor, and then illuminate all sections simultaneously, allowing us to characterize and correct for the relative differences in event timing and ensure proper synchronization between the camera and monitor. The measured event rates for the trigger pulse under different lighting conditions are shown in Fig.~\ref{supp:fig:supp_synch}.

\noindent \textbf{Camera/Monitor Synchronization Discrepancies:}
Even after accounting for the time and pixel synchronizations discussed in previous sections, we still observed a systematic timing bias in the monitor experiments, where the positive events excessively lagged behind the true star centroid as depicted in the third row and first column of Fig.~\ref{supp:fig:monitor_stars}. This error can be corrected by introducing a timing bias into the synchronization. Importantly, \textit{this timing bias is constant across all of the experiments}.  We determine this constant timing offset by what minimizes the average error for the maximum likelihood method (blue curve in Fig.~\ref{fig:results:monitor_centroiding} in the paper). Since maximum likelihood should provide the most accurate results, this approach ensures an optimal correction. While we acknowledge that this introduces some scientific bias, the primary concern in Fig.\ref{fig:results:monitor_centroiding} is the \textit{slope} (or standard deviation) of the centroiding error with respect to star magnitude. This is because centroiding inconsistency is the main source of error in the centroiding algorithms. The calculated timing offset does \textit{not} affect this slope, and both proposed methods—maximum likelihood estimation and the Gaussian approximation—exhibit significantly smaller slopes and standard deviations than alternatives. Finally, it is important to note that these method parameters were fitted to night sky measurements, not to the monitor data. This ensures that the parameters are not overfit to this specific monitor data.

% Note that we do not actually know the ground truth for the pulse timing; we just assume the``no filter'' response is instantaneous.

\noindent \textbf{Offset Curve:}
For the monitor centroiding experiments, we use the same offset curve as our real night sky data to ensure we are not overfitting to either modality. As shown in Fig.~\ref{fig:results:monitor_centroiding} of the paper, the offset curve significantly reduces centroiding error on the monitor, despite being calibrated on the night sky dataset.

\section{Real Night Sky Dataset Details}

\label{supp:additional-nightsky}

\indent \textbf{DVS Hardware:} Our event camera is one of the latest from Prophesee, the EVK4-HD.
The EVK4-HD consists of a 1280x720 event-pixel array (pixels are 4.86um in size) and related readout circuitry which is accessible over USB. 
The system is set up with a 35mm lens focused at infinity with a field of view of 5.7\textdegree x 10.2\textdegree. 
This leads to an effective instantaneous field of view for each pixel of approximately 30 arcseconds. 

\noindent \textbf{APS Hardware:} 
We utilize a Rocket Lab ST-16RT2 star tracker as our reference attitude-determination solution. 
The ST-16RT2 consists of a 2592 x 1944 pixel CMOS active-pixel sensor, a custom 16mm f/1.6 lens, a 12 cm rigid sun-keepout baffle, and an onboard processor capable of providing attitude solutions at 2Hz over an RS-485 link which we access via a USB to RS-485 converter. 
The manufacturer states the tracker can achieve accuracies of 5 arcseconds RMS cross-boresight and 55 arcseconds RMS around-boresight, with a maximum slew rate of 3\textdegree/second. 
The system is set up for a field of view of 15\textdegree x 20\textdegree.

\noindent \textbf{Sensor Stage Motion:}
We utilize two Edelkrone HeadONE units set up in a pan/tilt configuration to sweep the night sky with both the EVK4-HD and the ST-16RT2, which are rigidly mounted to each other.
These are controlled wirelessly via an Edelkrone Motor Controller dongle.
For each experiment, after a short static calibration period the pan/tilt mechanism is used to rotate both sensors at various speeds to simulate satellite rotation.

\begin{figure}
    \centering
    \includegraphics[width=\linewidth]{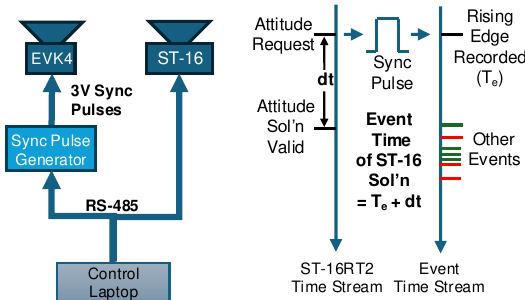}
    \caption{Left: Block diagram of synchronization hardware architecture. Right: Visualized timelines for the commercial star tracker (left) and event camera (right). Attitude requests come from the control laptop, which triggers the Sync Pulse Generator to send a synchronization pulse which the event camera records in its own timestream alongside positive and negative events. The commercial star tracker solution includes the validity time offset $dt$, measured from when the solution request was received. By combining the event-time of the sync pulse and the validity offset, the timestamp of the commercial star tracker's solution relative to the event stream can be recovered. }
    \label{fig:supp:sync_hardware_diagram}
\end{figure}
\noindent \textbf{DVS/APS Time Synchronization:}
We synchronize our event camera with the Rocket Lab star tracker to align our orientation estimates temporally as described in Fig.~\ref{fig:supp:sync_hardware_diagram}. 
To do so, we set up a small microcontroller (an Adafruit Trinket M0) which listens to the RS-485 serial line used to communicate with the Rocket Lab tracker. 
When a solution is requested by our control computer via this serial line, the microcontroller sends a square wave pulse to the EVK4 which records the pulse in the event stream with the appropriate timestamp.
Rocket Lab attitude solutions include a validity timestamp relative to the time of the initial solution request/square wave pulse which has been recorded in the EVK4 event stream. 
This combination of recorded data allows us to identify the precise time when the Rocket Lab-estimated attitude is valid in the event data timeline. 

\noindent \textbf{Relative Attitude Calibration:}
To accurately compare attitude solutions between the Rocket Lab APS tracker and the DVS camera, we need to determine the relative rotation between them. 
Assuming the cameras share a rigid body (i.e., are mounted to each other), the relative rotation remains constant regardless of the cameras' orientations. 
Therefore, we generally evaluate performance based on the deviation of the DVS tracker from this relative rotation. 
While stereo calibration using pixel values would be ideal, the Rocket Lab tracker provides quaternion solutions that are precisely calibrated to its imaging plane and undergo proprietary signal processing corrections inaccessible to the user. 
Consequently, we must rely directly on the quaternion solutions from the Rocket Lab tracker.

To obtain the relative rotation, we perform velocity sweeps of the night sky with both cameras, as shown in the velocity sweep experiments in Fig.~\ref{fig:experimental:track_profiles}. 
We then run our tracker over these sweeps and identify the rotation that minimizes the error between the Rocket Labs solutions and our tracker’s solutions. 
Mathematically, we determine the relative rotation $q_r$ that minimizes the rotation error between the two trackers:

\begin{align} \min_{\mathbf{q}_r}\sum_i||\mathbf{q}_{\text{RL}}^i \cdot \mathbf{q}_r \boxminus \mathbf{q}_{\text{DVS}}^i|| \end{align}

Here, the $\boxminus$ sign represents the boxminus operator for subtracting rotations, as described in Hertzberg et al.~\cite{hertzberg_integrating_2011}. 
Importantly, we observe that the choice of tracking algorithm has minimal impact on the estimated relative rotation, as any bias (i.e., centroiding error) is removed by performing the velocity sweeps in opposite directions. 
We measured the consistency of our relative rotation across three velocity sweeps to be within 20 arcseconds. 
We use the same relative rotation on all tracks and separate the data used for rotation calibration from test data to ensure there is no data leakage. 

\label{supp:sec:offcurve}
\noindent \textbf{Offset Curve Calibration:}
Knowledge of the cameras' relative rotation enables us to register events with the ``ground truth'' star locations predicted by the APS tracker. 
We measure the frequency of events occurring near star locations, generating the histograms like those shown in the middle row of Fig.~\ref{supp:fig:monitor_stars}. 
These histograms allow us to measure the offset of the event mean to the star's true location. 
The measured offset for each star magnitude results in our offset calibration curve. 
In our experiments, we used a velocity sweep pattern (which is \textit{not} used for evaluation in the main paper) to obtain the data for creation of the calibration curve.

\noindent \textbf{Computing the Theoretical Offset Curve:} To calculate the theoretical offset curve $z(m_s)$ from Eq.~\ref{eq:methods:gauss_approx_likelihood} in the paper, we first need to find the dark current value $I_0$ in paper Eq.~\ref{eq:methods:eventrate} and the cutoff frequency $f_c$ values $a,b$ in paper Eq.~\ref{eq:methods:fc_equation}. With the relative attitude calibration performed in the previous section, we can find the observed offset between the APS tracker (i.e. ground truth) and the centroids of our EBS tracker for each given star magnitude in the field of view. Given these offsets and also knowledge of the star size ($\sigma_s=2 \text{ pixels}$) and speed ($50 \text{ pix/s}$), we have an empirical offset curve. Substituting these values into paper Eqs.~\ref{eq:methods:eventrate},~\ref{eq:method:gauss_star}, \ref{eq:methods:fc_equation}, and \ref{eq:methods:complete_spatial_likelihood}, we then can optimize the values of $I_0, a, b$ such that the theoretical offset curve matches closely to the empirical curve, which lends the values $I_0=1$ (normalized so that a star with $m_s=7$ has peak intensity of $1$), $a=20 \text{ Hz/intensity}$, and $b = 2$ Hz. This curve is shown in Fig.~\ref{fig:methods:event_profile}c in the paper.

It is important to note three things about this calibration. The first is that we actually optimize the \textit{normalized} offset curves, where the minimum offset value $z(m_s)$ is set to 0. This is to remove systematic calibration offsets between our different modalities, and therefore we are matching the \textit{slope} of the $z(m_s)$ curve with respect to $m_s$, which is more important for accuracy calibration. The second is that we also test these same values of $I_0, a, b$ in the monitor experiments in Section \ref{sec:centroiding-results} in the paper and show near-perfect agreement. This suggests that we are not overfitting to the night-sky results but that these values do encompass the sensor characteristics. Finally, the offset curve theoretically does change slightly as a function of star speed $\mathbf{v}$. However, empirically in our night sky dataset we see that it hardly changes the offset, perhaps due to small synchronization issues or calibration. Therefore we just use the same offset curve for different star speeds. Future work should explore this relationship.

%Additionally, we use the velocity sweeps to derive the offset curves needed to correct for the offsets described in Section~\ref{sec:event-signal-model}. Specifically, we compute the event profile for stars of varying magnitudes, as illustrated in Figure~\ref{fig:methods:event_profile} and plot the offset of the event mean against the star magnitude as in Figure~\ref{fig:methods:event_profile}d. For our experiments, we generate the offset calibration curve using data from one track and apply this curve across all tracks.

\

\input{graphics/tables/dataset_table}

\section{Night Sky Dataset Discussion}
\label{supp:night-sky-dataset}

In comparison to other works (see Table~\ref{tab:supp:prior_datasets_comparison} for a summary of key dataset features), our dataset is unique in being collected from stars visible in an actual night sky.
Other datasets were collected using a monitor to simulate stars.
While simulating stars allows for reconstructing ground truth of the induced `orientation' of the event camera star tracker,  this type of data has downsides:
\begin{itemize}
    \item Monitor pixels are not perfectly uniform and may introduce unrealistic fixed pattern noise into event data collected from a monitor.
    \item The dynamic range of a monitor is limited and quantized to discrete pixel values.
    A monitor is therefore unable to represent any portions of a star's light which are dimmer than its smallest `on' value (i.e. 1 if the range of available values is 0-255) or brighter than its highest `on' value.
    A monitor star is also limited to a set of discrete brightness levels, while stars in the night sky can produce a continuous set of brightnesses. 
    \item Pixels in a monitor have a finite update rate and a quantized location (a monitor star cannot be located between pixels), introducing unrealistic event patterns into monitor data and sub-pixel inaccuracies in star location. 
    The night sky has an effectively infinite update rate, as we slew the sensors over unchanging light sources (stars). 
\end{itemize}

Tables~\ref{table:main-results} (of the main text) and \ref{tab:supp:prior_datasets_comparison} summarize and compare to other works the dataset we test on and will release for further study.
Most of our experiments are approximately 2.5 to 5 minutes long, ensuring we have a long enough time period to uncover possible drift in various trackers.
The total amount of data we release is significantly longer than prior datasets.
%Velocity Sweep experiments test tracker capabilities across a range of velocities, which includes much higher velocities than prior datasets.
%The Roll Sweep experiment is designed to thoroughly test about-bore performance, which tends to be worse than cross-bore.
%The Multipose experiments test a reasonable variety of direction changes at a constant velocity.

\noindent \textbf{Drift Phenomenon:} In some tracks, we observed that the mean difference between the trackers changed as they moved across the sky, with the change becoming more extreme near the horizon. An example of this ``drift'' in difference can be observed in the across plot of Fig.~\ref{fig:results:supp-ablations2}, where the difference gradually increases and decreases throughout the track. 
One potential explanation is a differential flexure of the cameras, which could change their relative rotation as they point closer to the horizon. 
In particular, the Rocket Lab tracker has a large sun keep-out baffle on it, which may lead it to be more deflected by gravity than the much smaller event camera when the optical axis is more perpendicular to the gravity vector. 
Another potential culprit is atmospheric diffraction towards the horizon impacting the accuracy of the trackers~\cite{jannin2024calibration}. Since the two sensor (APS and EBS) are not perfectly optically aligned, they may experience different attitude errors due to atmospheric refraction. In one experiment, we collected data in a small range of angles near the zenith, and observed that the relative error between the APS and EBS attitude remained relatively constant (i.e., we did not observe the drift phenomenon). Since tracks measured closer to zenith (further from the horizon) are less susceptible to the drift effect, they have a 10-20 second lower arcsecond difference on average.

\section{Additional Results}
\label{supp:additional-tracking-results}

\begin{figure*}
  \centering
  \includegraphics[width=1\linewidth, page=1]{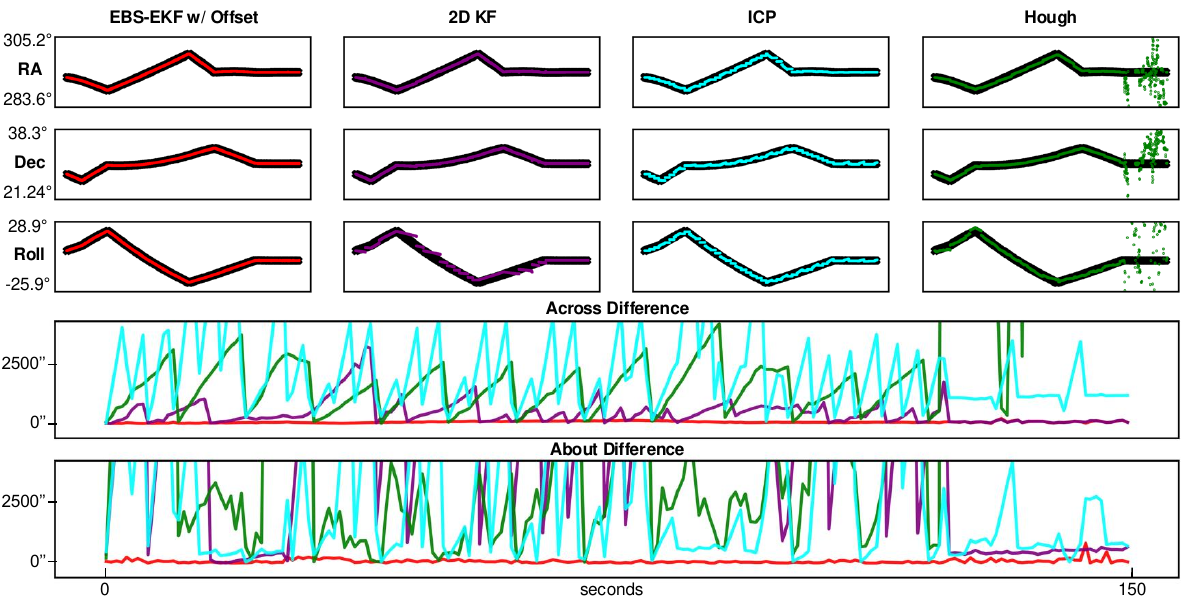}
   \caption{\textbf{Multipose 1}. RL samples are shown in black and each method is shown in a unique color.}
   \label{fig:results:supp-errors1}
\end{figure*}

\begin{figure*}
  \centering
\includegraphics[width=1\linewidth, page=2]{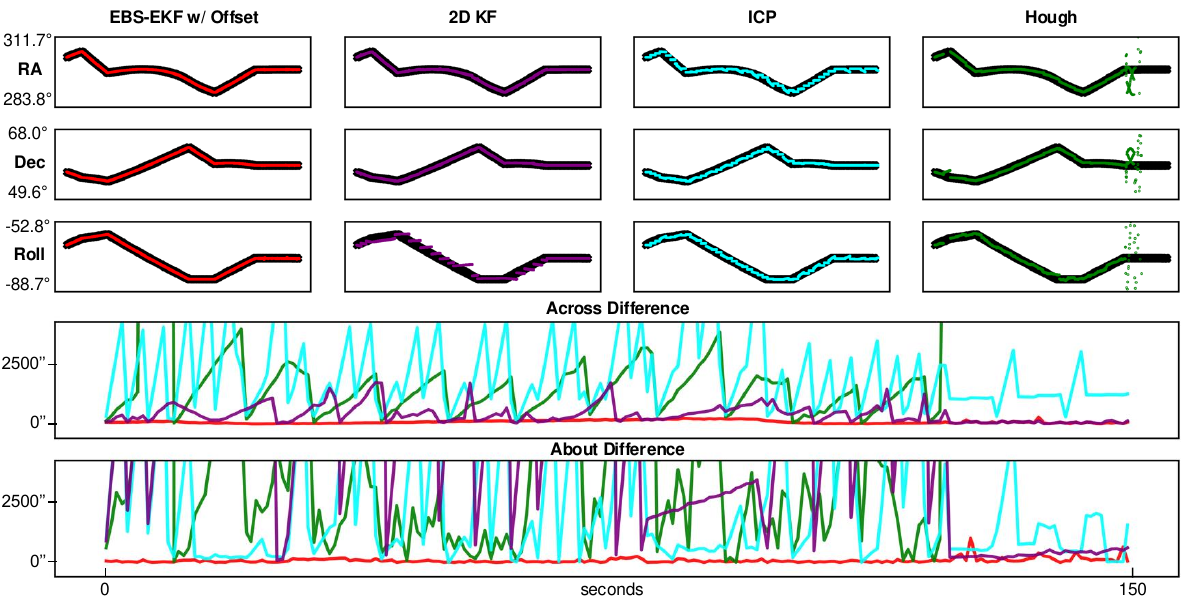}
   \caption{\textbf{Multipose 2}. RL samples are shown in black and each method is shown in a unique color.}
   \label{fig:results:supp-errors2}
\end{figure*}

\begin{figure*}
  \centering
\includegraphics[width=1\linewidth, page=5]{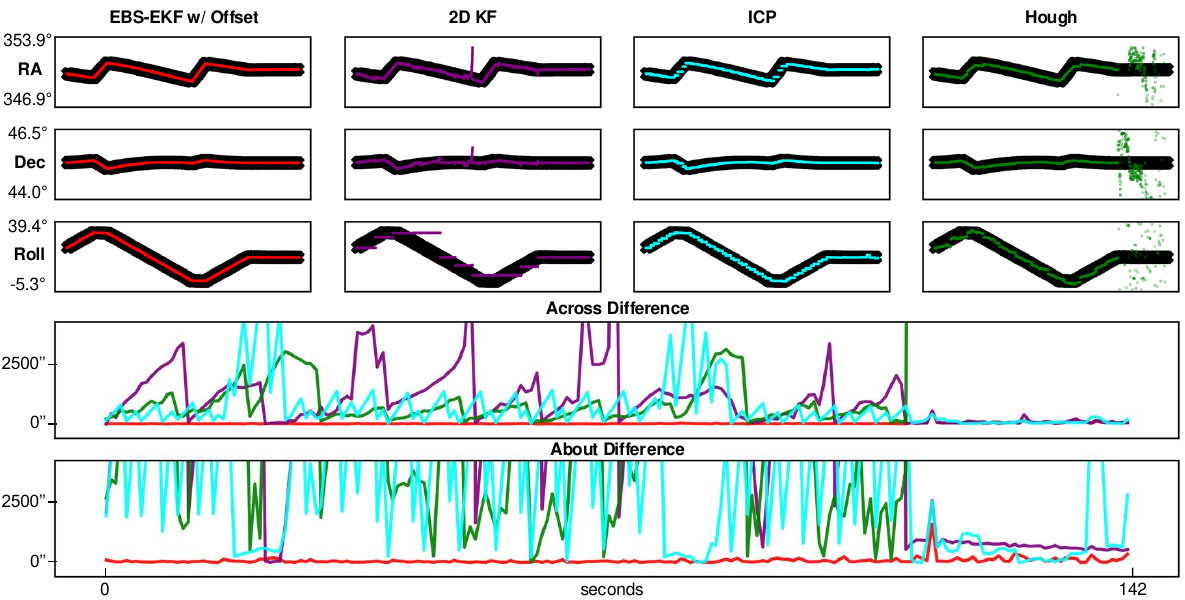}
   \caption{\textbf{Multipose 3}. RL samples are shown in black and each method is shown in a unique color.}
   \label{fig:results:supp-errors3}
\end{figure*}

\begin{figure*}
  \centering
\includegraphics[width=1\linewidth, page=6]{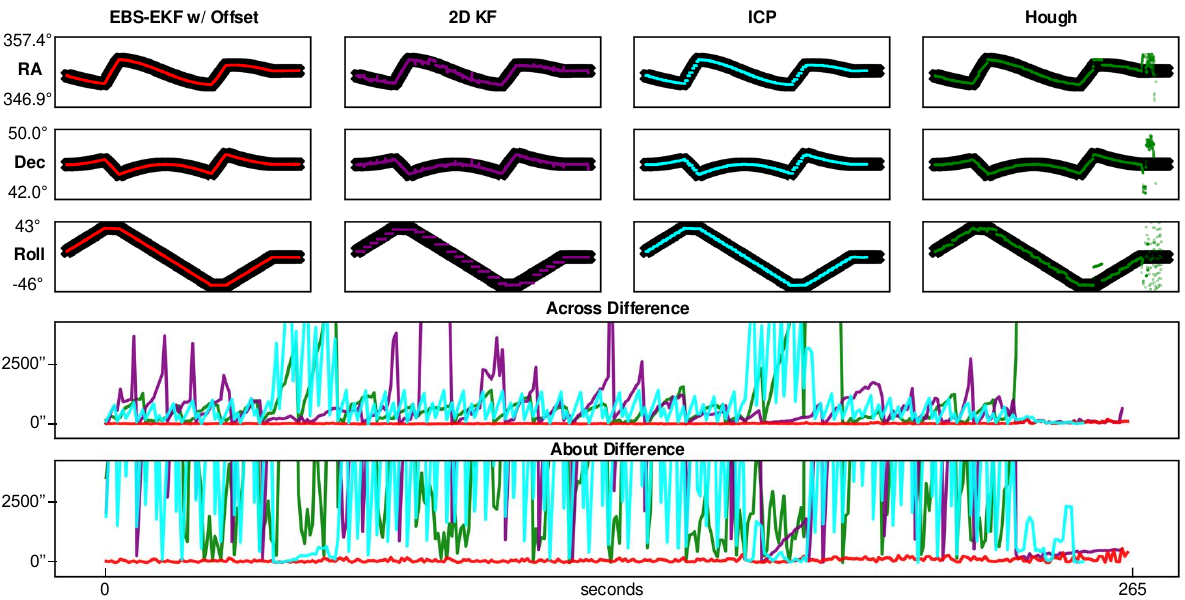}
   \caption{\textbf{Multipose 4}. RL samples are shown in black and each method is shown in a unique color.}
   \label{fig:results:supp-errors4}
\end{figure*}

\begin{figure*}
  \centering
\includegraphics[width=1\linewidth, page=3]{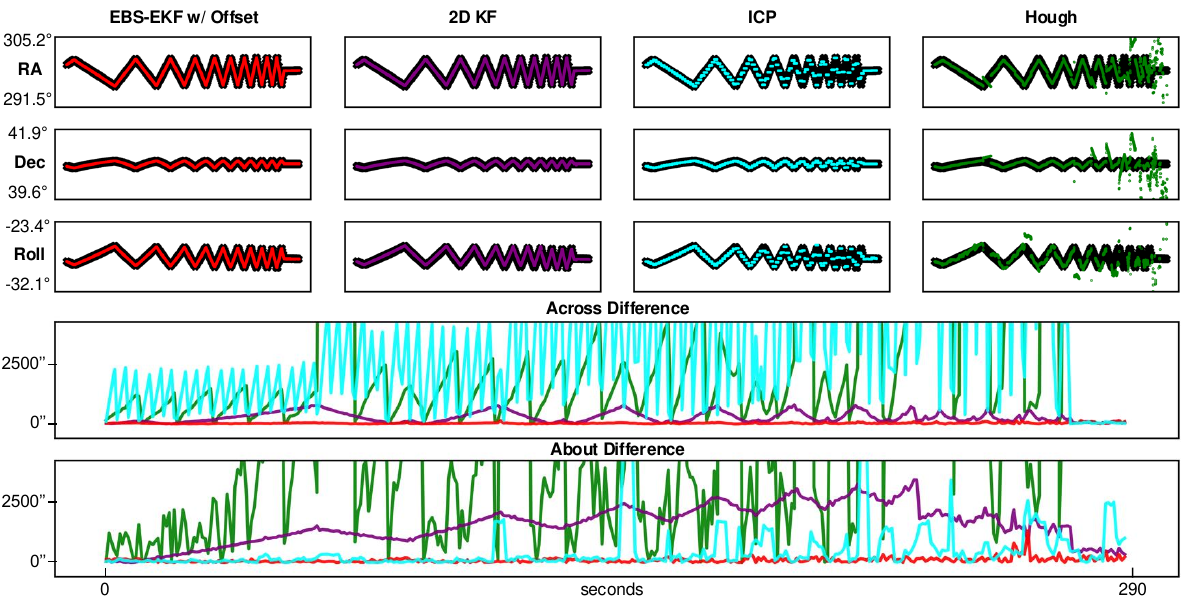}
   \caption{\textbf{Velocity Sweep 1}. RL samples are shown in black and each method is shown in a unique color.}
   \label{fig:results:supp-errors5}
\end{figure*}

\begin{figure*}
  \centering
\includegraphics[width=1\linewidth, page=4]{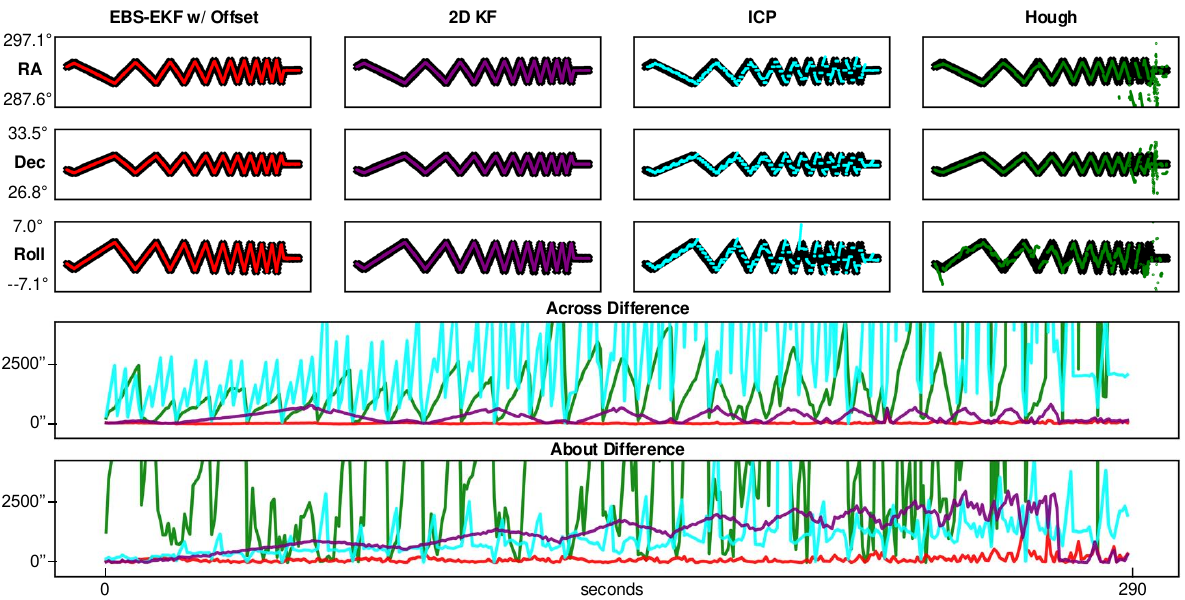}
   \caption{\textbf{Velocity Sweep 2}. RL samples are shown in black and each method is shown in a unique color.}
   \label{fig:results:supp-errors6}
\end{figure*}

\begin{figure*}
  \centering
\includegraphics[width=1\linewidth, page=7]{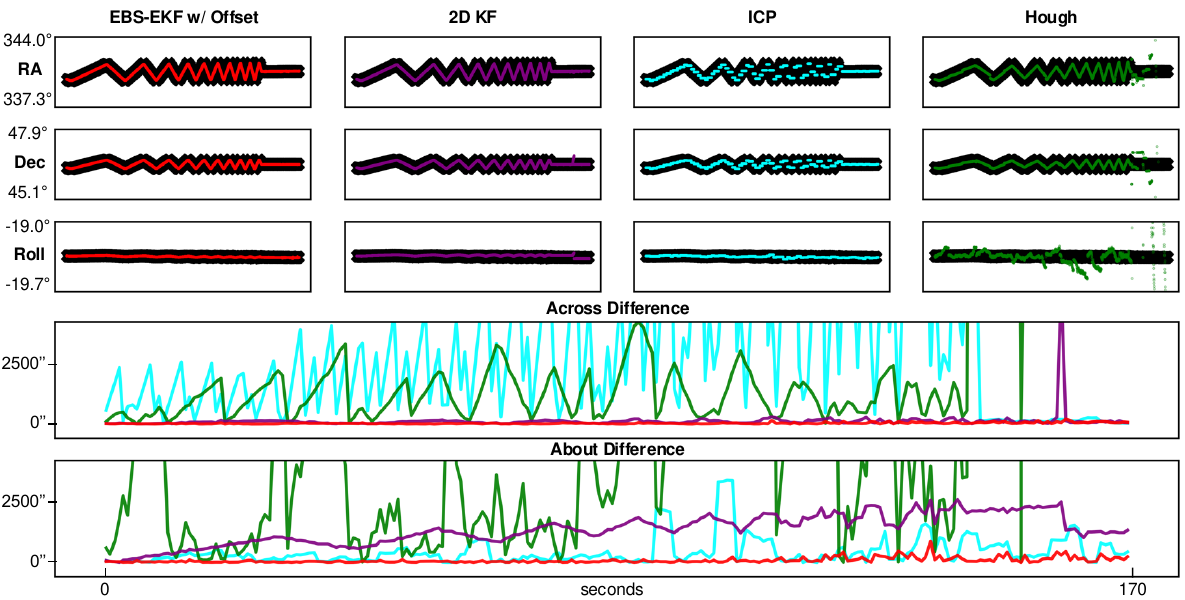}
   \caption{\textbf{Velocity Sweep 3}. RL samples are shown in black and each method is shown in a unique color.}
   \label{fig:results:supp-errors7}
\end{figure*}

\begin{figure*}
  \centering
\includegraphics[width=1\linewidth, page=8]{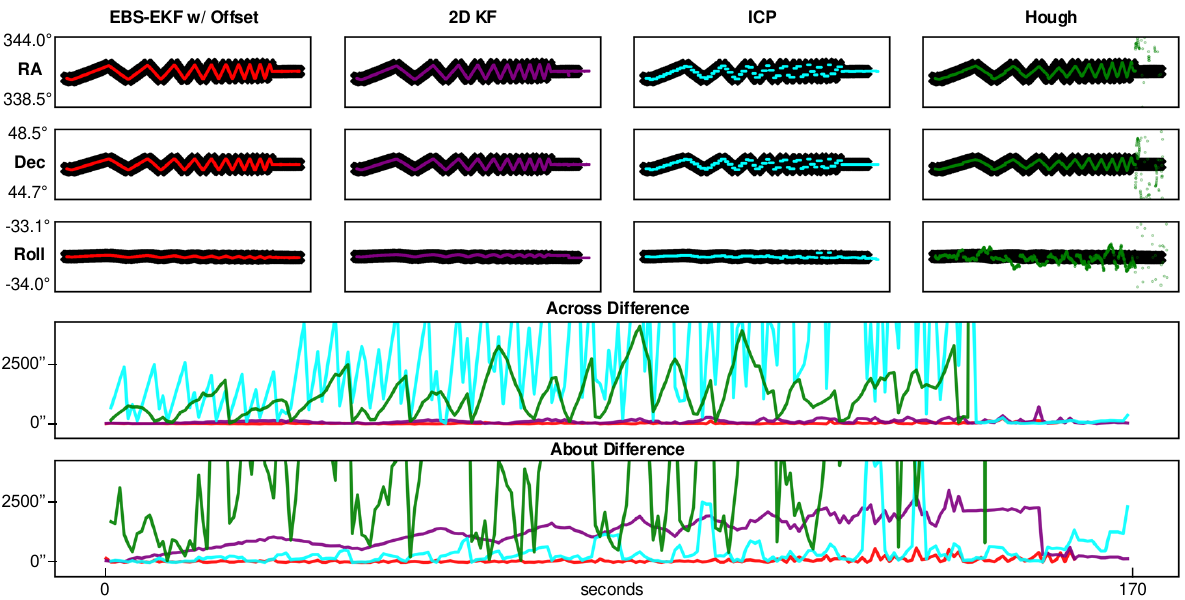}
   \caption{\textbf{Velocity Sweep 4}. RL samples are shown in black and each method is shown in a unique color.}
   \label{fig:results:supp-errors8}
\end{figure*}

\begin{figure*}
  \centering
\includegraphics[width=1\linewidth, page=9]{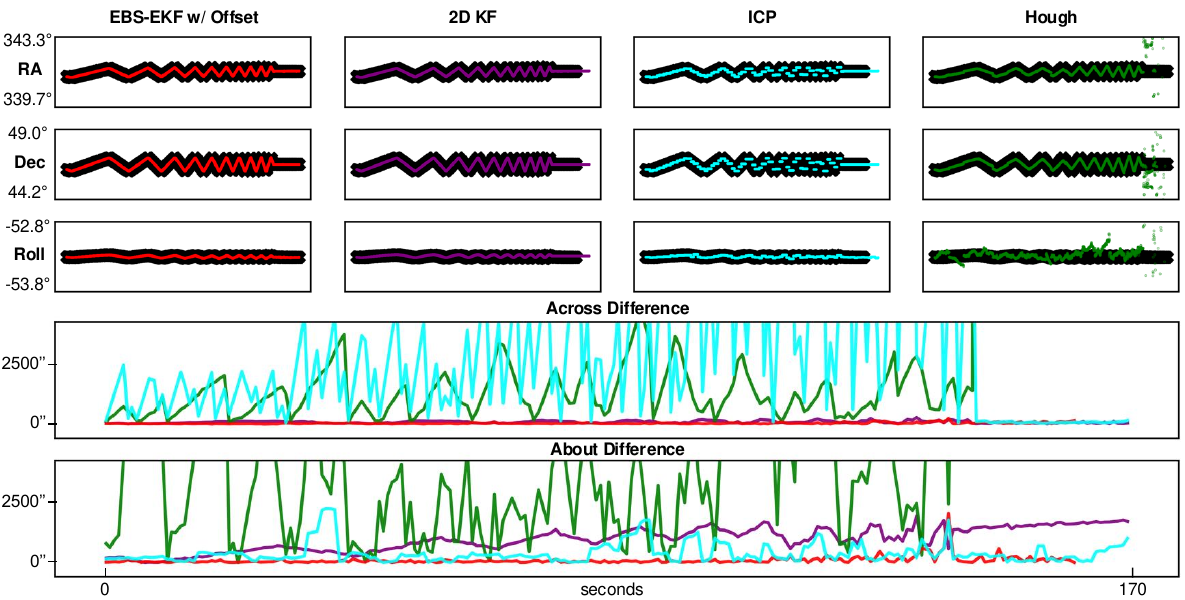}
   \caption{\textbf{Velocity Sweep 5}. RL samples are shown in black and each method is shown in a unique color.}
   \label{fig:results:supp-errors9}
\end{figure*}

\begin{figure*}
  \centering
\includegraphics[width=1\linewidth, page=10]{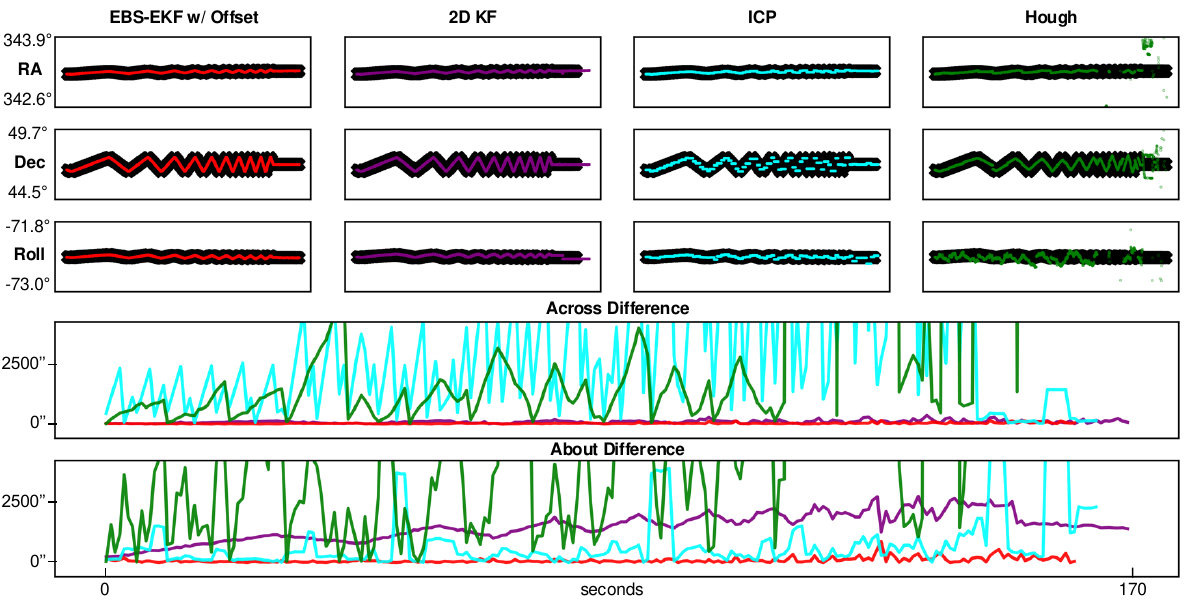}
   \caption{\textbf{Velocity Sweep 6}. RL samples are shown in black and each method is shown in a unique color.}
   \label{fig:results:supp-errors10}
\end{figure*}

\begin{figure*}
  \centering
\includegraphics[width=1\linewidth, page=13]{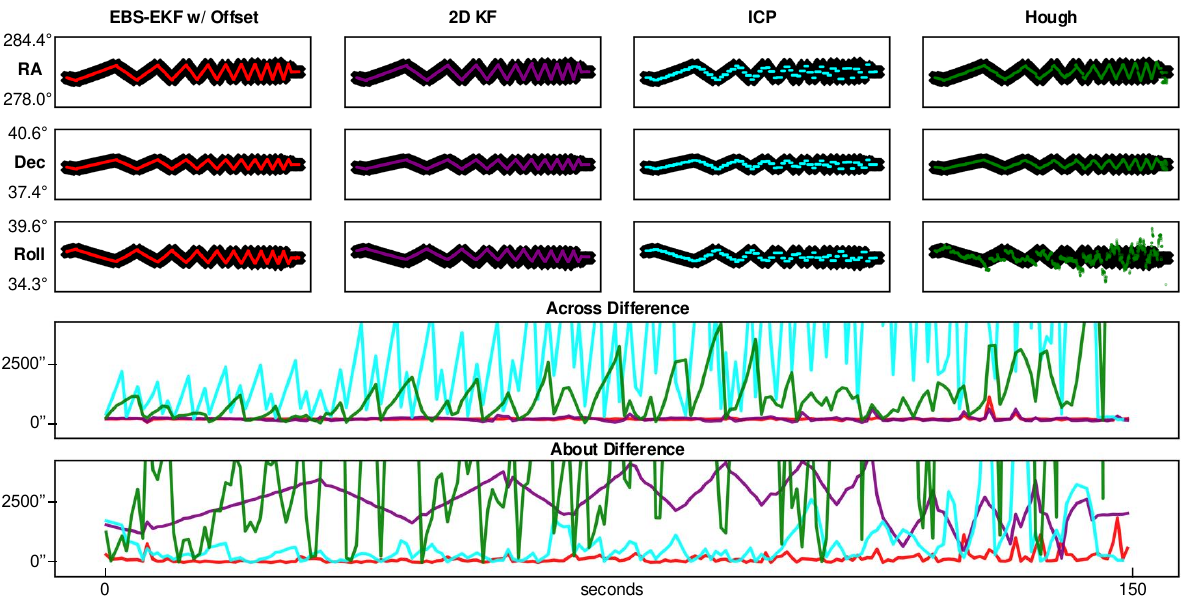}
   \caption{\textbf{Velocity Sweep 7}. RL samples are shown in black and each method is shown in a unique color.}
   \label{fig:results:supp-errors11}
\end{figure*}

\begin{figure*}
  \centering
\includegraphics[width=1\linewidth, page=11]{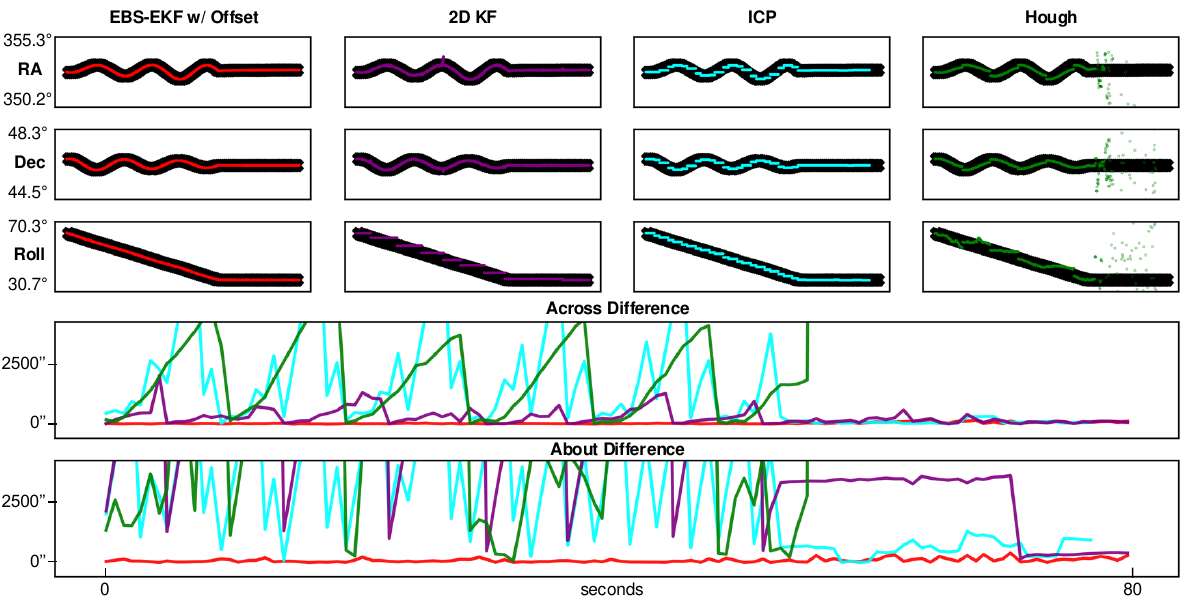}
   \caption{\textbf{Smooth Sine}. RL samples are shown in black and each method is shown in a unique color.}
   \label{fig:results:supp-errors12}
\end{figure*}

\begin{figure*}
  \centering
\includegraphics[width=1\linewidth, page=12]{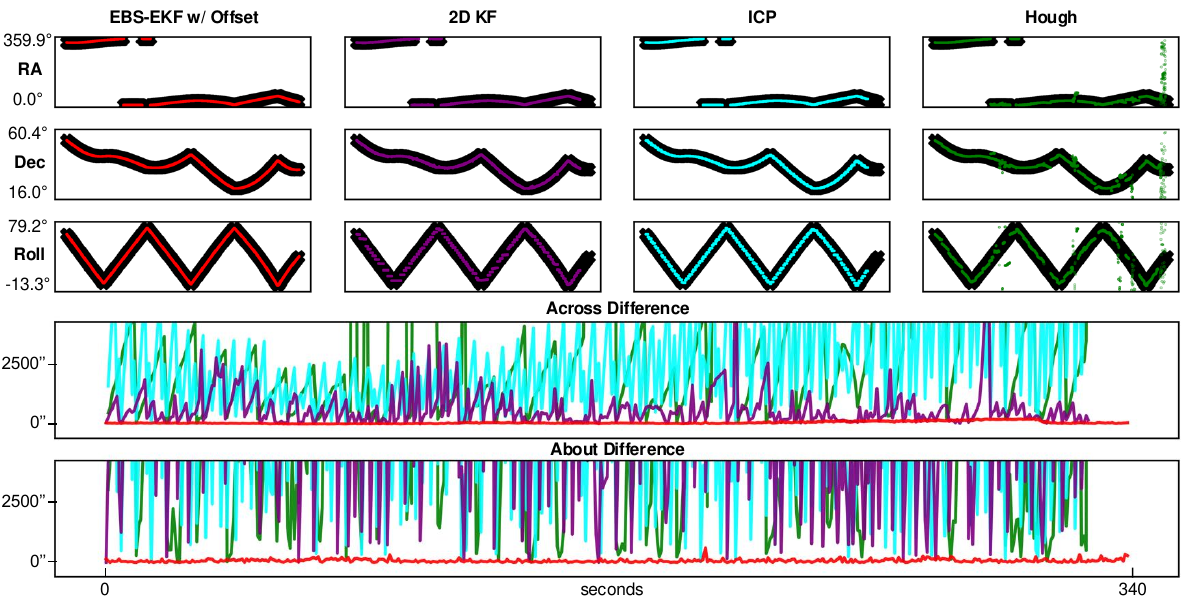}
   \caption{\textbf{Tilt Ladder}. RL samples are shown in black and each method is shown in a unique color.}
   \label{fig:results:supp-errors13}
\end{figure*}

\begin{figure*}
  \centering
\includegraphics[width=1\linewidth, page=3]{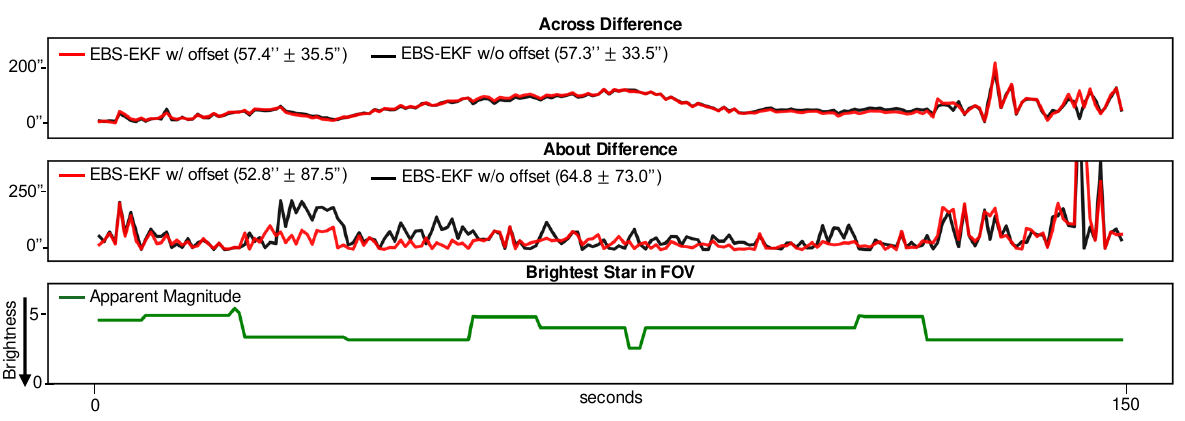}
   \caption{\textbf{Multipose 1 EBS-EKF w/ offset vs EBS-EKF w/o Offset}}
   \label{fig:results:supp-ablations1}
\end{figure*}

\begin{figure*}
  \centering
\includegraphics[width=1\linewidth, page=4]{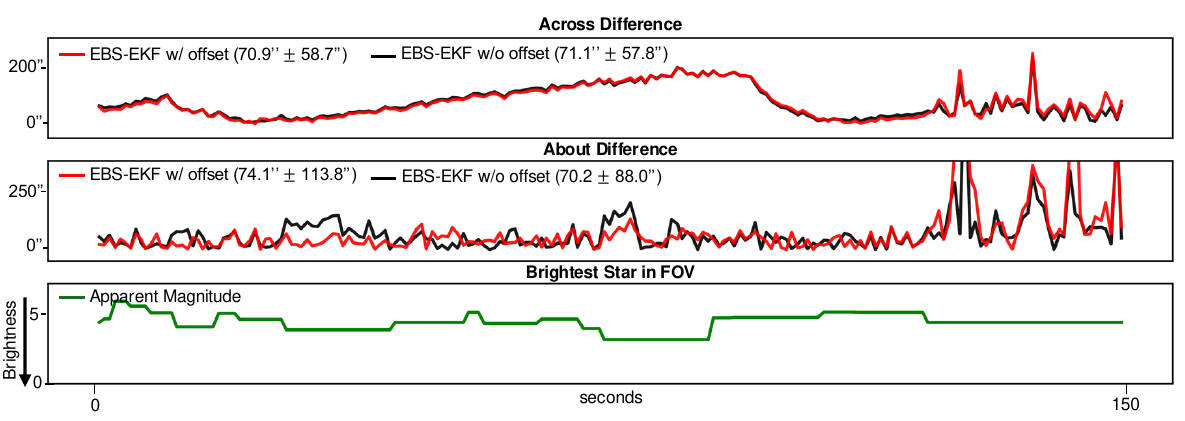}
   \caption{\textbf{Multipose 2 EBS-EKF w/ offset vs EBS-EKF w/o Offset}}
   \label{fig:results:supp-ablations2}
\end{figure*}

\begin{figure*}
  \centering
\includegraphics[width=1\linewidth, page=5]{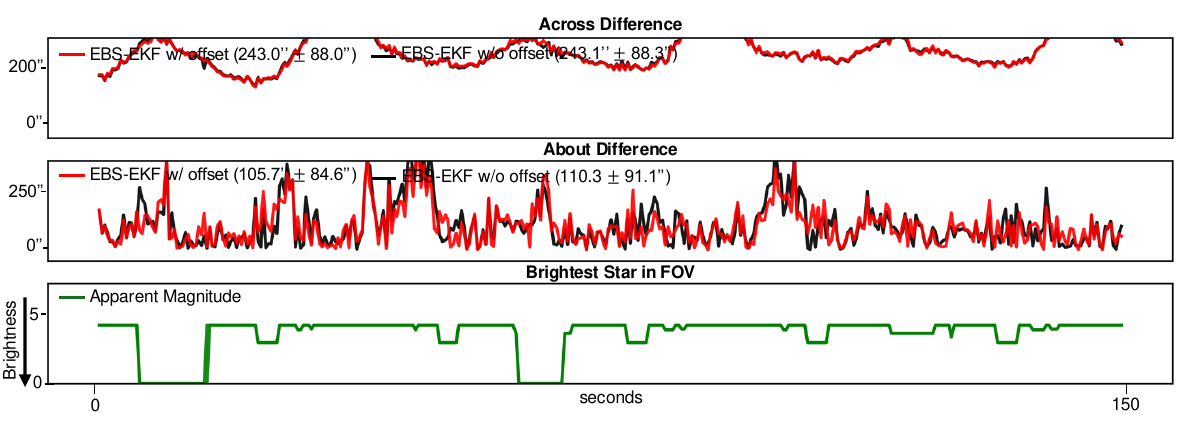}
   \caption{\textbf{Multipose 3 EBS-EKF w/ offset vs EBS-EKF w/o Offset}}
   \label{fig:results:supp-ablations3}
\end{figure*}

\begin{figure*}
  \centering
\includegraphics[width=1\linewidth, page=1]{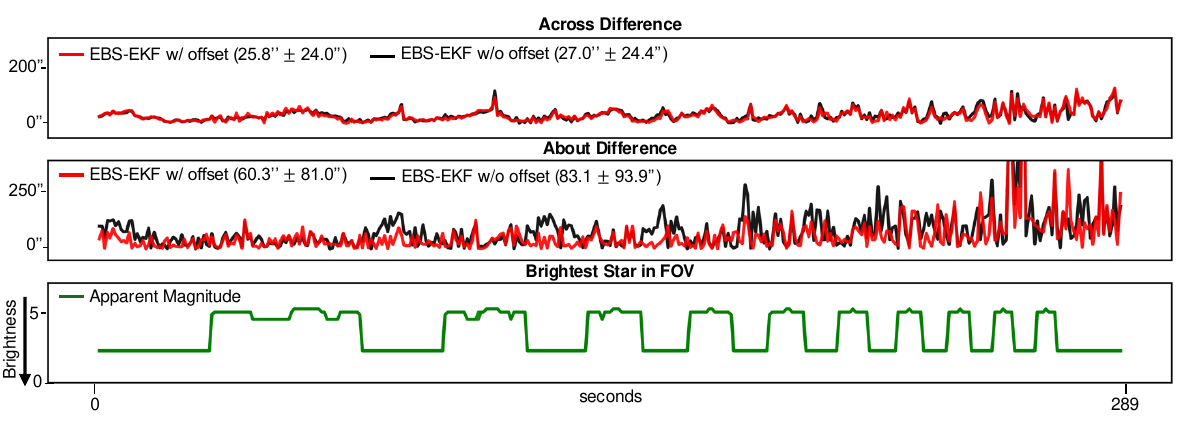}
   \caption{\textbf{Velocity Sweep 1 EBS-EKF w/ offset vs EBS-EKF w/o Offset}}
   \label{fig:results:supp-ablations4}
\end{figure*}

\begin{figure*}
  \centering
\includegraphics[width=1\linewidth, page=2]{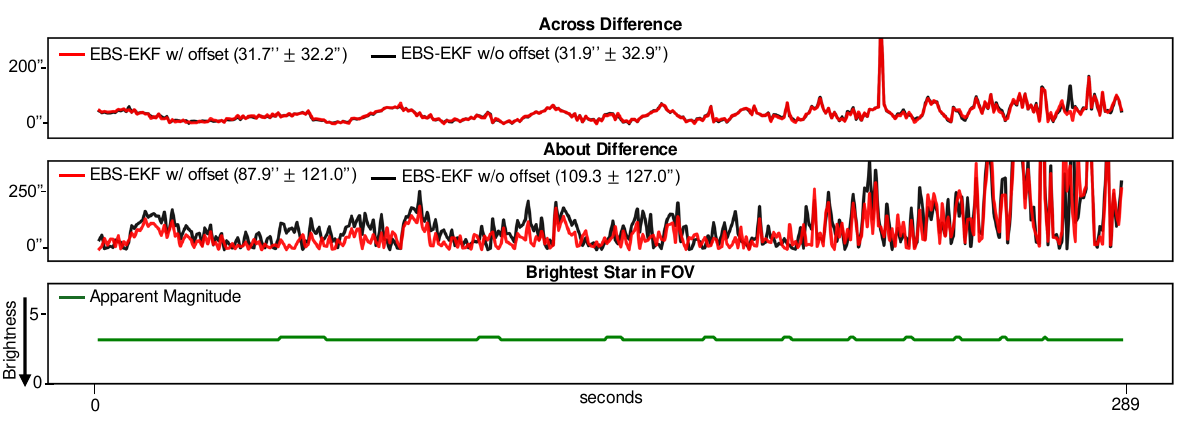}
   \caption{\textbf{Velocity Sweep 2 EBS-EKF w/ offset vs EBS-EKF w/o Offset}}
   \label{fig:results:supp-ablations5}
\end{figure*}

\begin{figure*}
  \centering
\includegraphics[width=1\linewidth, page=9]{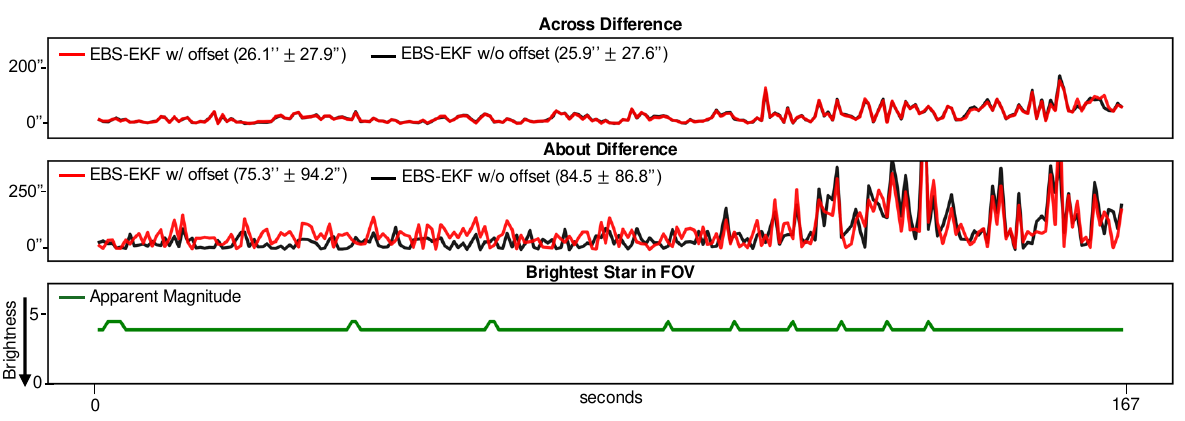}
   \caption{\textbf{Velocity Sweep 3 EBS-EKF w/ offset vs EBS-EKF w/o Offset}}
   \label{fig:results:supp-ablations6}
\end{figure*}

\begin{figure*}
  \centering
\includegraphics[width=1\linewidth, page=10]{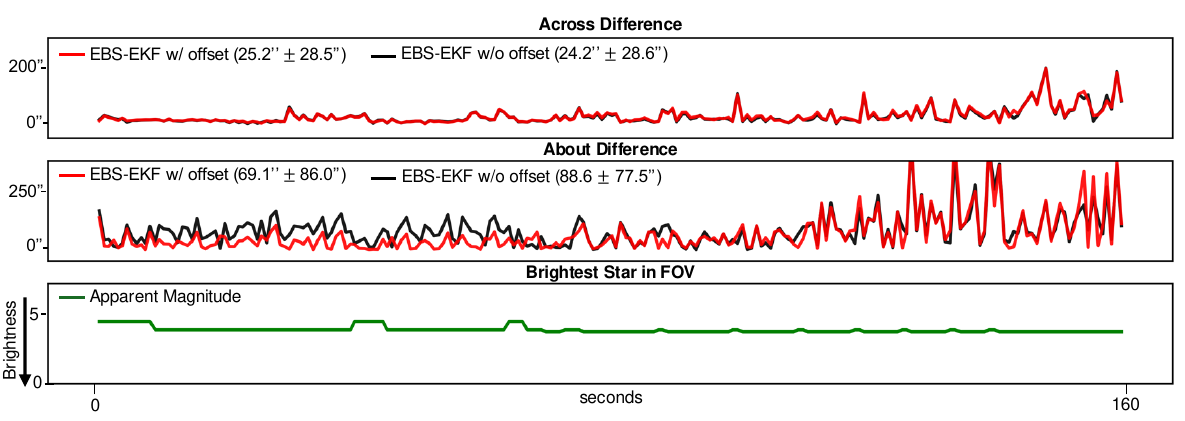}
   \caption{\textbf{Velocity Sweep 4 EBS-EKF w/ offset vs EBS-EKF w/o Offset}}
   \label{fig:results:supp-ablations7}
\end{figure*}

\begin{figure*}
  \centering
\includegraphics[width=1\linewidth, page=11]{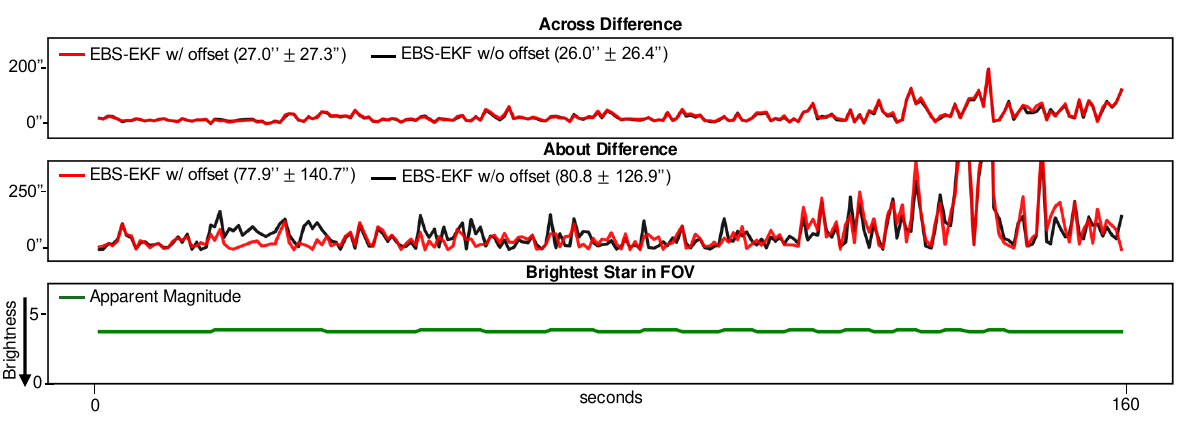}
   \caption{\textbf{Velocity Sweep 5 EBS-EKF w/ offset vs EBS-EKF w/o Offset}}
   \label{fig:results:supp-ablations8}
\end{figure*}

\begin{figure*}
  \centering
\includegraphics[width=1\linewidth, page=12]{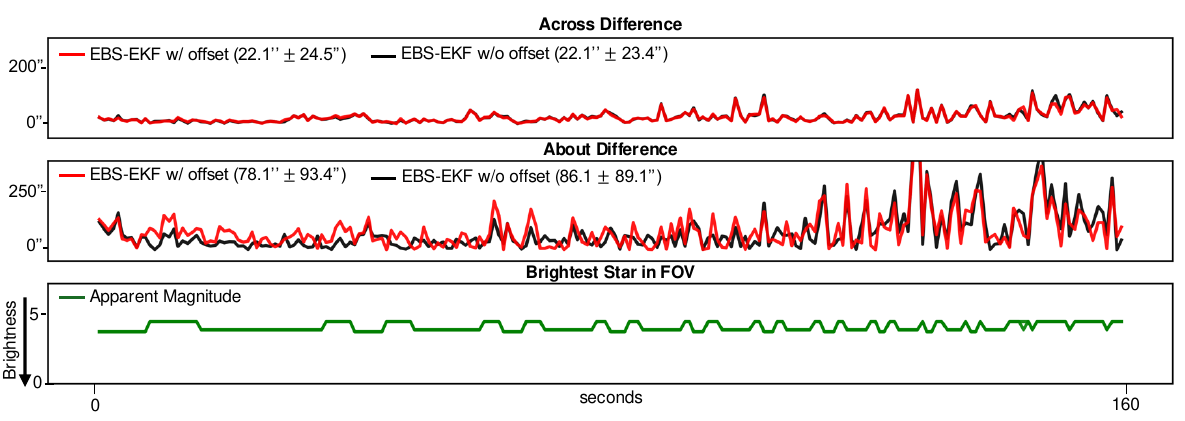}
   \caption{\textbf{Velocity Sweep 6 EBS-EKF w/ offset vs EBS-EKF w/o Offset}}
   \label{fig:results:supp-ablations9}
\end{figure*}

\begin{figure*}
  \centering
\includegraphics[width=1\linewidth, page=13]{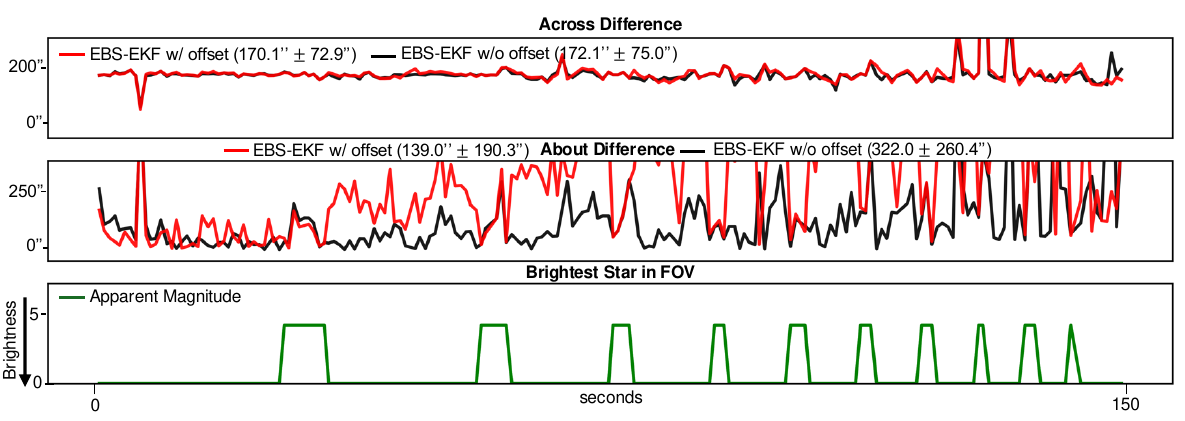}
   \caption{\textbf{Velocity Sweep 7 EBS-EKF w/ offset vs EBS-EKF w/o Offset}}
   \label{fig:results:supp-ablations10}
\end{figure*}

\begin{figure*}
  \centering
\includegraphics[width=1\linewidth, page=8]{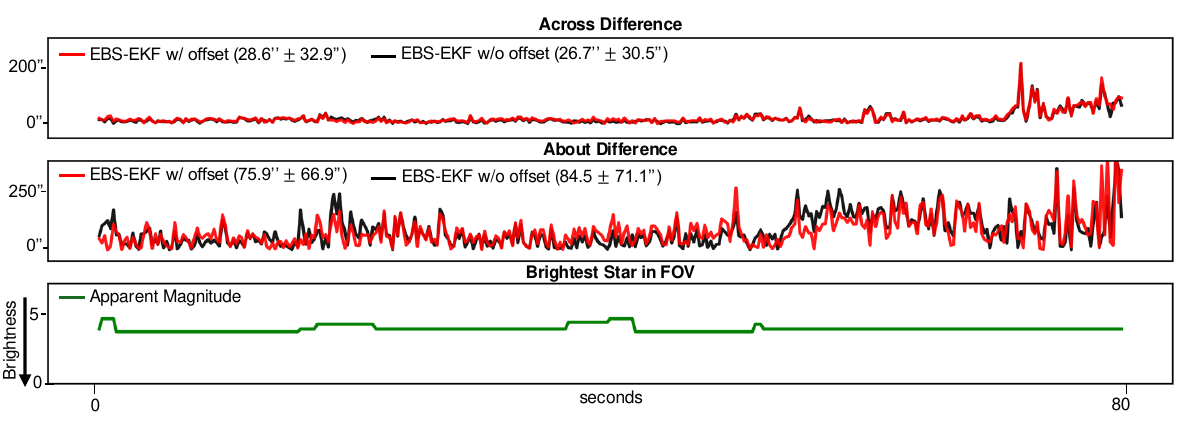}
210   \caption{\textbf{Smooth Sine EBS-EKF w/ offset vs EBS-EKF w/o Offset}}
   \label{fig:results:supp-ablations11}
\end{figure*}

\begin{figure*}
  \centering
\includegraphics[width=1\linewidth, page=14]{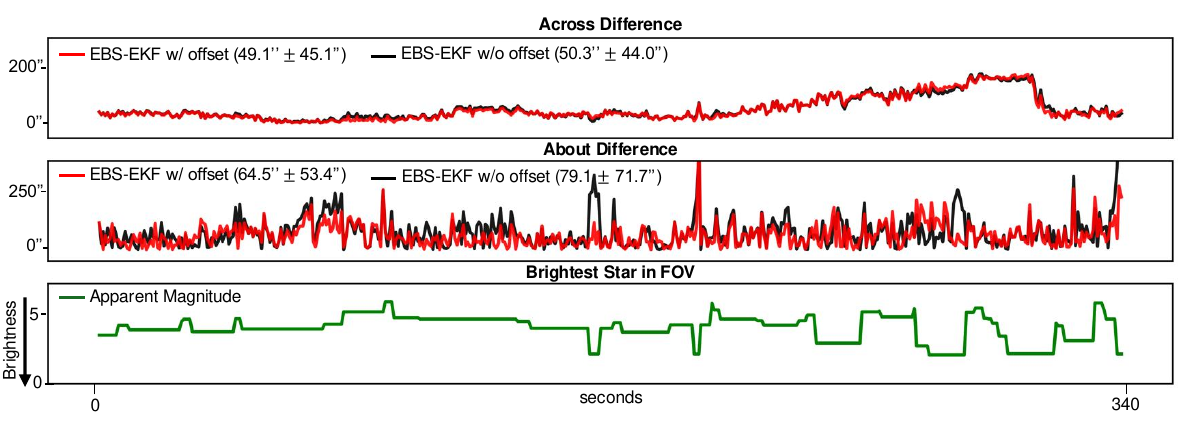}
   \caption{\textbf{Tilt Ladder EBS-EKF w/ offset vs EBS-EKF w/o Offset}}
   \label{fig:results:supp-ablations12}
\end{figure*}

\noindent \textbf{Additional Plots:} We provide difference plots (Figs.~\ref{fig:results:supp-errors1} -~\ref{fig:results:supp-errors13}) for all tracks shown in Table~\ref{table:main-results} of the main text, including plots comparing performance with and without our offset correction (Figs.~\ref{fig:results:supp-ablations1} -~\ref{fig:results:supp-ablations12}). Referring to the difference plots, our method is typically an order-of-magnitude more accurate than the existing methods. Referring to the offset correction plots, we observe that using our offset typically reduces the error by 30 arcseconds, particularly in the roll direction, which is highly sensitive to centroiding errors~\cite{liebe_accuracy_2002}. We observe that the offset can drastically improve performance when a bright star is in the FOV. For instance, Fig.~\ref{fig:results:supp-ablations10} shows results for a track that passed over Vega, an exceptionally bright star with an apparent magnitude of .03. In this case, our offset correction improves the mean average difference by over 180 arcseconds.  

\noindent \textbf{Video Results: } We include three video results in the supplemental folder. In \textit{high-velocity-track.mp4}, we display event frames for the high velocity track shown in Fig.~\ref{fig:teaser} of the main text. White circles circle events that were used to update the tracker's attitude (i.e., white circles indicate stars). Note that while we display events as frames (using 30 ms batch times), the algorithm operated asynchronously at 1 KHz; the events are shown as frames for visualization purposes. The video \textit{high\_vel.mp4} shows the camera frustum for the high velocity track. We show the reconstructed track by our algorithm in red and the APS track in purple. The APS track failed to provide solutions in the fast regions of the track, resulting in an incorrect track interpolation. Our method provides an accurate reconstruction of the track. We also include a visualization of the frustum for the velocity sweep track in \textit{vel\_sweep\_1.mp4}.

% \section{Code Release}
%\label{supp:code-release}

%We have Python implementations of our method and existing methods, as well as processing for all of our datasets. We will release this code upon the acceptance of the paper. 

%%%% ABLATIONS %%%%

%\section{Additional Camera Geometry}
%Given a set of $m$ star centroids on the imaging plane $\mathbf{x}_{i:m}$, absolute or relative tracking methods find a set of corresponding stars in world coordinates $\mathbf{s}_{i:m} \in \mathbf{S}$. Then, the camera's attitude can be recovered by solving Wahba's problem~\cite{wahba1965least},
%\begin{align}
%\mathbf{q_{\text{wc}}} = \argmin_{\hat{\mathbf{q}} \in SO(3)} \sum_i^m ||\vv{\mathbf{x}_i} - \hat{\mathbf{q}}(\mathbf{s}_i)||_2^2
%\label{eq:wahba}
%\end{align}
%which finds the rotation that minimizes the squared error between the backprojected points $\vv{\mathbf{x}_i}$ and the star world coordinates. The backprojected points are given by
%\begin{align}
%    \vv{\mathbf{x}} = \frac{\mathbf{K}^{-1}\bar{\mathbf{x}}}{||\mathbf{K}^{-1}\bar{\mathbf{x}}||_2}
%\end{align}
%where $\mathbf{K} \in \real^{3 \times 3}$ is the camera intrinsic matrix and $\bar{\mathbf{x}} = [\mathbf{x}^T, 1]^T$ represents the pixel in homogeneous coordinates.  Singular value decomposition may be used to find the root mean squared deviation solution to Wahba's problem~\cite{horn1987closed}, which we use to compute absolute measurements. However, our primary tracking mode estimates attitude asynchronously using an Extended Kalman Filter.

%% file: graphics/tables/dataset_table.tex
\begin{table*}[t]

\small
\centering
\begin{tabular}{@{}lcccccc@{}}
\toprule
\multicolumn{1}{c}{Dataset} &
  Camera &
  Resolution &
  FOV &
  \begin{tabular}[c]{@{}c@{}}Max \\ Slew Rate\end{tabular} &
  \begin{tabular}[c]{@{}c@{}}Total\\  Length\end{tabular} &
  Stars \\ \midrule
\multicolumn{1}{l|}{\begin{tabular}[c]{@{}l@{}}Chin/\\ Bagchi\end{tabular}} &
  \begin{tabular}[c]{@{}c@{}}Davis \\ 240C\end{tabular} &
  \begin{tabular}[c]{@{}c@{}}240\\ x180\end{tabular} &
  \begin{tabular}[c]{@{}c@{}}20 \\ deg\end{tabular} &
  \begin{tabular}[c]{@{}c@{}}4 \\ deg/s\end{tabular} &
  \begin{tabular}[c]{@{}c@{}}4.5 \\ min\end{tabular} &
  LCD \\
\multicolumn{1}{l|}{Latif} &
  \begin{tabular}[c]{@{}c@{}}Proph.\\  EVK3\end{tabular} &
  \begin{tabular}[c]{@{}c@{}}1280\\ x720\end{tabular} &
  \begin{tabular}[c]{@{}c@{}}1.5\\  deg\end{tabular} &
  \begin{tabular}[c]{@{}c@{}}0.005 \\ deg/s\end{tabular} &
  \begin{tabular}[c]{@{}c@{}}70 \\ sec\end{tabular} &
  LCD \\
\multicolumn{1}{l|}{\textbf{Ours}} &
  \textbf{\begin{tabular}[c]{@{}c@{}}Proph.\\  EVK4\end{tabular}} &
  \textbf{\begin{tabular}[c]{@{}c@{}}1280\\ x720\end{tabular}} &
  \textbf{\begin{tabular}[c]{@{}c@{}}10\\  deg\end{tabular}} &
  \textbf{\begin{tabular}[c]{@{}c@{}}7.5 \\ deg/s\end{tabular}} &
  \textbf{\begin{tabular}[c]{@{}c@{}}19.5 \\ min\end{tabular}} &
  \textbf{\begin{tabular}[c]{@{}c@{}}Night \\ Sky\end{tabular}}
\end{tabular}
\caption{EBS star tracking dataset comparisons. Our dataset contains more data and faster slew rates than prior works, in addition to be collected from an actual night sky instead of simulated stars on a monitor.}

\label{tab:supp:prior_datasets_comparison}
\end{table*}